\documentclass[3p,a4paper]{elsarticle}
\usepackage[english]{babel}
\usepackage[utf8]{inputenc}
\usepackage[cmex10]{amsmath}
\usepackage{lineno,hyperref}
\usepackage{xcolor}
\usepackage{graphicx}
\usepackage{array}
\usepackage{mdwmath}
\usepackage{mdwtab}
\usepackage{eqparbox}
\usepackage{pdflscape}
\usepackage{pbox}
\usepackage{color}
\usepackage{nccmath}
\usepackage{amssymb}
\usepackage{multirow}
\usepackage{longtable}
\usepackage{array,multirow,graphicx}
\usepackage{subfig}
\usepackage{lscape}
\usepackage{setspace}
\usepackage{todonotes}
\usepackage{longtable}
\modulolinenumbers[5]
\usepackage{enumitem}
\journal{International-journal-of-approximation-reasoning}

\begin{document}

\begin{frontmatter}

\title{Time Series Forecasting Using Fuzzy Cognitive Maps: A Survey}

\author[minds]{Omid Orang\corref{mycorrespondingauthor}}
   \cortext[mycorrespondingauthor]{Corresponding author}
   \ead{omid.orang2009@gmail.com}
\author[ifnmg,minds]{Petr\^onio C\^andido de Lima e Silva}
   \ead{petronio.candido@ifnmg.edu.br}
\author[dee,minds]{Frederico Gadelha Guimar\~aes}
   \ead{fredericoguimaraes@ufmg.br}
   \ead[url]{https://minds.eng.ufmg.br/}
 
\address[minds]{Machine Intelligence and Data Science (MINDS) Laboratory, Federal University of Minas Gerais, Belo Horizonte, Brazil}
\address[ifnmg]{Federal Institute of Education Science and Technology of Northern Minas Gerais, Janu\'aria Campus, Brazil}
\address[dee]{Department of Electrical Engineering, Universidade Federal de Minas Gerais, Belo Horizonte, Brazil}

\begin{abstract}
Among various soft computing approaches for time series forecasting, Fuzzy Cognitive Maps (FCM) have shown remarkable results as a tool to model and analyze the dynamics of complex systems.
FCM have similarities to recurrent neural networks and can be classified as a neuro-fuzzy method. In other words, FCMs are a mixture of fuzzy logic, neural network, and expert system aspects, which act as a powerful tool for simulating and studying the dynamic behavior of complex systems. The most interesting features are knowledge interpretability, dynamic characteristics and  learning capability. The goal of this survey paper is mainly to present an overview on the most relevant and recent FCM-based time series forecasting models proposed in the literature. In addition, this article considers an introduction on the fundamentals of FCM model and learning methodologies. Also, this survey provides some ideas for future research to enhance the capabilities of FCM in order to cover some challenges in the real-world experiments such as handling non-stationary data and scalability issues.  Moreover, equipping FCMs with fast learning algorithms is one of the major concerns in this area.
\begin{singlespace}
\noindent 

\end{singlespace}

\end{abstract}


\end{frontmatter}


\section{Introduction}
\label{sec:introduction}

In recent decades, tackling real complex problems in a highly reliable way has become one of the major challenges for researchers. Increasing complexity comes from some factors including uncertainty, ambiguity, inconsistency, multiple dimensionalities, increasing the number of effective factors and relation between them. Some of these features are common among most real-world problems which are considered complex and dynamic problems. In other words, since the data and relations in real world applications are usually highly complex and inaccurate, modeling real complex systems based on observed data is a challenging task especially for large scale, inaccurate and non stationary datasets. Therefore, to cover and address these difficulties, the existence of a computational system with the capability of extracting knowledge from the complex system with the ability to simulate its behavior is essential. In other words, it is needed to find a robust approach and solution to handle real complex problems in an easy and meaningful way \cite{ye2015learning}.

Hard computing methods depend on quantitative values with expensive solutions and lack of ability to represent the problem in real life due to some uncertainties. In contrast, soft computing approaches act as alternative tools to deal with the reasoning of complex problems \cite{Zadeh1994association}. Using soft computing methods such as fuzzy logic, neural network, genetic algorithms or a combination of these allows achieving robustness, tractable and more practical solutions. Generally, two types of methods are used for analyzing and modeling dynamic systems including quantitative and qualitative approaches. In some cases, modeling complex and nonlinear systems through quantitative techniques is difficult and costly \cite{Aguilar2005surveyfcm}. In contrast, qualitative methods do not suffer from the mentioned restrictions. Fuzzy Cognitive Map (FCM) is a kind of important qualitative soft computing technique proposed by Kosko \cite{Kosko1986FCM} in 1986, as an extension of a cognitive map. FCM has attracted great attention among researchers, with high capability of solving dynamic and complex problems.

FCM is a graph knowledge-based method and like any traditional cognitive maps, composed of concepts and causal relationships among the concepts. The difference is that in FCMs the concepts are modeled by fuzzy sets and fuzzy connections define the relationship among them. In other words, the existence of the fuzzy feedback in  FCMs structure creates an option to extract and model the causal knowledge.  Based on  \cite{Papageorgiou2014FuzzyCM}, FCM can be described by two characteristics. The first one is the type of relationship among the concepts with different intensities represented by uncertain fuzzy numbers. The second one is the system dynamicity, that is, it evolves with time. In FCMs structures composed of feedbacks,  changes in one concept may influence other concepts. They have the ability to incorporate human knowledge and adapt it through learning procedures as the major advantage \cite{Parsopoulos2002pso}.

FCMs seem to have similarities to recurrent neural networks and can be classified as a neuro-fuzzy method. In other words, FCMs are a mixture of fuzzy logic, neural network, and expert system aspects, which act as a powerful tool for simulating and studying the dynamic behavior of complex systems. Therefore, FCMs can learn due to their dynamic structure from an Artificial Intelligence (AI) point of view. This learning capability of FCM improves its structure and computational behavior \cite{Parsopoulos2002pso,Papageorgiou2011ruleextraction,Jose2009routledge}. By considering this feature, diverse methods of learning algorithms have been proposed in the literature to improve the performance of the FCM,  such as Differential Hebbian Learning (DHL) \cite{Dickerson1994}, Genetic Algorithm \cite{froelich2009predictive,stach2005genetic}, Ant Colony Optimization \cite{ding2011first} etc.

Qualitative modeling, ease of perception, high ability to dealing with uncertainties, capability to represent nonlinear and causal behaviors, flexibility and explainability can be accounted as interesting attributes of FCMs \cite{Vliet2010LinkingSA}. According to these properties, plenty of papers were published in the field of FCMs that cover diverse domains involving decision making, control systems, time series forecasting, classification, electrical and software engineering, medicine etc \cite{stach2005surveyFCM,felix2019review,Papageorgiou2014FuzzyCM}. 

The goal of this paper is mainly to present an overview on the progress in the studies of time series forecasting utilizing FCMs. In other words, this paper provides an up-to-date survey covering some relevant proposed FCM-based time series forecasting techniques which are presented in Section \ref{sec:TSF using FCMs}. Before that, some major characteristics of FCMs are reminded. Thus, the remainder of this work is organized as follows: in Section \ref{sec:Fundamentals}, in Section \ref{sec:learning}, the most relevant learning strategies are introduced; Section \ref{sec:TSF using FCMs} focuses on a thorough review of some developments and updated FCM-based time series forecasting methods and finally Section \ref{sec:challenges} highlights some challenges and future research possibilities.

\section{Fuzzy Cognitive Maps}
\label{sec:Fundamentals}

\subsection{Fuzzy Cognitive Maps Fundamentals}
Fuzzy Cognitive Map (FCM) proposed by Kosko \cite{Kosko1986} is a particular family of Cognitive Map (CM) theory that was introduced by Axelrod \cite{Axelrod1976}. The term of CM was introduced to investigate the cognitive activities of rats using some learning experiments on their choice of an appropriate path to food. Later Axelrod, a political scientist, used the knowledge of people to form the CM in the form of causal relationships between concepts to formulate decision making in political science.

CM is a causal and qualitative model composed of concepts (nodes) and the causal connection among the concepts. These directed signed arrows among the concepts are represented as weights that reflect the effect of one node on another one. These assigned directions among the concepts of CM can be positive or negative. In \citep{eden1992,EDEN2004673} several methodologies have been suggested to represent the CM and analyze its complexity. If the node influences on others, we say it is a cause and if influenced by others, it is an effect (cause and effect causality). In binary cognitive maps (BCM) the concept labels are mapped to binary states denoted as $a_i \in \{0, 1\}$, where the value 1 means that the concept is activated. Also, the weights belong to the crisp set $w_{ij} \in \{-1, 0, 1\}$ in BCM \citep{Axelrod1976}. The value $1$ represents, positive causality, that the activation (change from $0$ to $1$) of concept $c_{i}$ occurs concurrently with the same activation of concept $c_{j}$ or that deactivation (change from $1$ to $0$) $c_i$ occurs concurrently with the same deactivation of concept $c_{j}$. The value $-1$ represents the opposite situation, in which the activation of $c_{i}$ deactivates the concepts $c_{j}$ or vice versa. The $w_{ij} = 0$ means that there are no concurrently occurring changes of the states of the concepts. CMs were used in various areas such as decision-making \citep{EDEN2004673,Eden2004colin,Jonathan1982CM}, financial \citep{Dong2004cm} and other areas. 

Accordingly, FCM is a hybrid soft computing technique based on the CM concept as a combination of fuzzy logic and CM to present uncertainties and complex characteristics of the systems. FCM has also emerged as a powerful paradigm for knowledge representation by providing a flexible mechanism for knowledge representation of intelligent systems \citep{Miao2001Dynamicalcognitivenetwork,Papakostas2011TrainingFC,salmeron2009FCMLMS,taber1991knowledge,Taber2007QuantizationEO}. In FCM there exist partial causal relations among the concepts. Thus these connections can be represented using fuzzy subsets which indicate vague values in the range $[-1,1]$. It means that the elements of the weight matrix in FCM can be full positive (+1), full negative (-1), or any value positioned in the interval $[-1,1]$. Thus, this feature discriminates FCM from CM which enables FCM to deal with imprecise and uncertain data. For each FCM with k number of concepts, the casual relation among concepts is represented by one $k\times k$ square weight matrix (connection matrix) ${W}$.
\begin{equation}\label{eq:W}
{W}=\left(
\begin{array}{ccc}
w_{11}&\ldots& w_{1k}\\
\vdots&\ddots &\vdots\\
w_{k1}&\ldots& w_{kk}
\end{array}\right)
\end{equation}

Each member $w_{ij}$ of the matrix denotes the directed connection between concepts $c_i$ and $c_j$ which governs the intensity of the relationships between a pair of concepts to measure their influence degree on each other. On the other hand, the strength of causal relations among concepts is represented by absolute fuzzy value. The higher the absolute value of weight among concepts leads to  the stronger (closer) relationship among couple of concepts. In details, there exist three possible options for each individual of the weight matrix as follows:

\begin{enumerate}
    \item \textbf{${w}_{ij}<0$}: means negative connection among concepts $c_i$ and $c_j$. It means that a decrease/increase in the value of node $c_i $ makes an increase/decrease in the value of node $c_j$.
    \item \textbf{${w}_{ij}>0$}: means positive connection among concept $c_i$ and $c_j$. It means that a decrease/increase in the node $c_i$  makes a decrease/increase in the node $c_j$. 
    \item \textbf{${w}_{ij}=0$}: confirms null relation among nodes $c_i$ and $c_j$.
    \end{enumerate}
    
\begin{figure}[htb]
    \centering
    \includegraphics[width=1\textwidth]{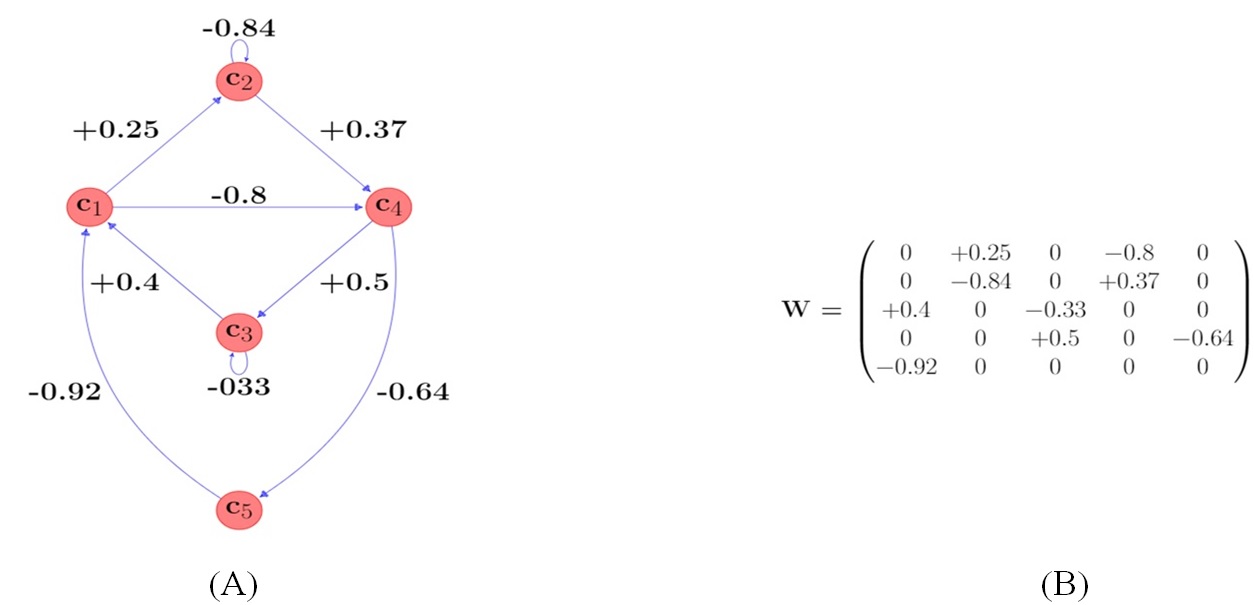}
    \caption{Simple FCM with 5 nodes (A) graphical structure (B) Weight matrix}
    \label{fig:FCMrepresentation}
\end{figure}

Figure \ref{fig:FCMrepresentation}-A shows a simple example of FCM with 5 concepts while Figure \ref{fig:FCMrepresentation}-B presents the connections among the concepts, as weight matrices, with the dimension of $5\times 5$.

From the graphical perspective, FCM is composed of concepts and connections among them that are represented by signed, weighted, and directed arcs. So, FCM is a signed, weighted digraph. In other words, the degree, sign, and direction of the influence of one concept on the others are required, apart from the number of concepts. ${C} = [c_1,\ldots,c_k ]$ is the set of $k$ concepts which are variables (nodes of the graph) that compose the system and $W$ is the connection matrix among the concepts. FCMs are defined by the 4-tuple  $(C, W, a,f)$ where the third element,  $a = ({a_1}, \ldots,{a_k} )$ is the state vector containing the set of all concepts. At any time $t$, each concept $c_i$ has an activation or fuzzy value restricted into the interval $[0,1]$. On the other hand, FCM operation is based on the application of a specific inference rule which calculates the next value of each concept by considering the total impacts of the concepts directly connected to this one in each iteration.  Generally, based on the literature \citep{papakostas2010classifying,felix2019review}, there exist different kinds of inference rules with respect to the past concept’s value and self-connection relation for each concept. Equation \eqref{eq:activation_rule1} indicates Kosko’s transition rule for concept $c_i$ at time $t+1$ with no self-connection and without memory, which is widely used in many FCM-based applications.

\begin{equation}\label{eq:activation_rule1}
{a_i}(t+1)=f\left( \sum_{j=1,j\neq i}^{k} w_{ji} {a_j}(t) \right)
\end{equation}
where $k$ is the number of concepts and ${w}_{ij}$ highlights the causal connection among concept $c_i$ and $c_j$, whereas ${a_i}(t)$ denotes the sate value of concept $c_i$ at time step $t$. It can be said that ${a_i}(t+1)$ and ${a_i}(t)$  are response vector and state vector of FCM respectively. Once the values of all weights of the FCM (i.e. the relationship matrix of FCM) are determined, FCM starts from a given initial state vector to carry out reasoning through the consecutive iterations computation according to the equation \eqref{eq:activation_rule1}. At each time step, the FCM generates a state vector that contains all concept activations. The state vector ${a}(t) = ({a_1}(t), \ldots, {a_k}(t) )$ refers to the activation level of all concepts at time $t$. This updating rule is iteratively repeated until the termination condition is satisfied. It means that in any iteration a new state value of the concepts is gained and after a certain number of repetitions, one of the three possible states can occur for FCM \cite{Kosko1986} including fixed equilibrium point, limited cycle and chaotic behavior. In both cases of chaotic or cyclic, the output may be partially unreliable because of the lack of stability. When the FCM reaches a fixed equilibrium point, it can be concluded that the FCM has been converged.

Although the mentioned updating rule has been used in many FCM-based applications, other modifications have been developed in the literature \cite{papakostas2010classifying,felix2019review}.

The investigation proposed in  \cite{Stylios2004complex,Parsopoulos2002pso} introduces the modified version of rule updating as the below formula indicates. In this way, in addition to the corresponding weights and activation values from other concepts, concepts consider their own past activation value known as memory factors for the nodes. Therefore, equation \eqref{eq:activation_rule2} shows Kosko’s updating rule with memory and without self-connection.

\begin{equation}\label{eq:activation_rule2}
{a_i}(t+1)=f\left( {a_i}(t)+\sum_{j=1,j\neq i}^{k} w_{ji} {a_j}(t) \right)
\end{equation}

Although the above equations ignore the effect of self-connection, the following one highlights the inference rule by adding self-connection.

\begin{equation}\label{eq:activation_rule3}
{a_i}(t+1)=f\left( {a_i}(t)+\sum_{j=1}^{k} w_{ji} {a_j}(t) \right)
\end{equation}

The other type of rule updating is described by taking only the effect of self-connection into account and ignoring the memory element.

\begin{equation}\label{eq:activation_rule4}
{a_i}(t+1)=f\left( \sum_{j=1}^{k} w_{ji} {a_j}(t) \right)
\end{equation}

Researchers in \cite{PAPAGEORGIOU201228} reported another version of updating inference rule without considering  memory and self connection according to the following equation that outperforms equation \eqref{eq:activation_rule1} when used for predictions tasks implemented on real-world data.

\begin{equation}\label{eq:activation_rule5}
{a_i}(t+1)=f\left( \sum_{j=1,j\neq i}^{k} w_{ji} (2{a_j}(t)-1))\right)
\end{equation}

The other alternative was proposed in \cite{Papageorgiou2011ruleextraction} to avoid the conflicts emerging in the case of inactive concepts. From this perspective, it can be said that the element $(2{a_i}(t)-1)$ acts like bias weights. More clearly, it prevents saturation problems where there is no information about the initial concepts state \cite{felix2019review}.

\begin{equation}\label{eq:activation_rule6}
{a_i}(t+1)=f\left( \sum_{j=1,j\neq i}^{k} w_{ji} (2{a_j}(t)-1)+(2{a_i}(t)-1)\right)
\end{equation}

Choosing the convenient updating rule to model the dynamic behavior of the system depends on the problem and a strong understanding of the target simulated system. 

The function $f$ in updating rule equations specifies the transfer/activation/limiting function used to maintain the activation state of each concept to the predefined range. Table \ref{tab:fcm_activation} highlights the most commonly used activation functions in the literature \cite{felix2019review}. Both bivalent and trivalent are discrete and generate a finite number of states. The hyperbolic tangent and the sigmoid function are categorized in the continuous group and produce infinite states. Unlike generating finite states by using the first two discrete functions, hyperbolic tangent and sigmoid functions as continuous functions, generate an infinite number of states used for simulating qualitative and quantitative cases. Nevertheless, the studies verified the higher performance of FCM by applying continuous activation functions \cite{Tsadiras2008AFCMactivations}. A benchmarking study was organized in \cite{bueno2009benchmarking} where the best performance is obtained using the sigmoid activation function in comparison to the other ones.

In sigmoid activation function $\gamma$ is steepness parameter (function slope). The higher values of steepness value, the more sensitive to the changes of $x$.  Depending on the various problems properties, different amounts of $\gamma$ can be tuned. Generally, the values of $\gamma$ are selected up to five in most of the references in the literature. For instance, in \cite{Aguilar2005surveyfcm} the value of $\gamma$ has been adopted to $5$. Furthermore, a dynamical optimization of $\gamma$ has been suggested in \cite{Salmeron2016dynamicoptimisation,oikonomou2013particle} to choose the optimum value according to the problem definition.

\begin{table}[]
    \centering
\begin{tabular}{|c|c|} \hline
\textbf{Activation Function}    & \textbf{Mathematical Representation} \\ \hline
bivalent   & 
$f(x) = \left\{ \begin{array}{ccc}
        0 & if & x <0  \\
        1 & if & x \geq 0
    \end{array}\right.$ \\ \hline
trivalent   & 
$ f(x) = \left\{ \begin{array}{ccc}
        1 & if & x \geq 0.5  \\
        0 & if & -0.5 \leq x \leq 0.5  \\
       -1 & if & x\leq -0.5 
        
    \end{array}\right.$ \\ \hline

 hyperbolic tangent  & 
$ f(x) = \frac{\exp^{2x}-1}{\exp^{2x}+1} $\\ \hline

sigmoid  & 

$ f(x) = \frac{1}{\exp^{-\gamma {x}}+1}$ \\ \hline
\end{tabular}
\caption{Most widely used FCM activation functions }
    \label{tab:fcm_activation}
\end{table}

\subsection{High Order Fuzzy Cognitive Maps (HFCM)}
Dynamically speaking, equations \eqref{eq:activation_rule1} to \eqref{eq:activation_rule5} express the first order FCM. It means that through this way only the first-order dynamics of FCM will be obtained. Rewriting them, for instance equation \eqref{eq:activation_rule4}, indicates precisely that the activation value of each concept, the degree to which the observed value of a time series ${y_i}(t)$ belongs to the fuzzy set, at a particular time step $(t+1)$, only relies on the activation level of all concepts at $t$ moment, as displayed in the following formula:
\begin{equation}\label{eq:rewritten_eq3}
 {a}(t+1)=f({W} \cdot {a}(t) )
\end{equation}
where ${a}(t) = [{a_1}(t),{a_2}(t),\dots, {a_k}(t)]$ indicates the vector of the activation value of all nodes in time $t$. Similarly, ${a}(t+1) = [{a_1}(t+1),{a_2}(t+1),\dots, {a_k}(t+1)]$ denotes the vector of the activation level of all nodes in time $t+1$. Furthermore, $W$ is the weight matrix that stores the edge values among the concepts. 

Due to the aforementioned constraint, an accurate modeling of FCM will not be captured only by considering the current activation value of the concepts and ignoring the past values.  For the purpose of tackling the mentioned restriction and elevate the FCMs performance, especially for describing the dynamic behavior of the complex systems more accurately, high order FCMs was introduced in which the reasoning rule equation is modified and described as follows \cite{Shanchao2018WHFCM,LIU2020106105EMDHFCM,Lu2014HFCMCMEANS}:

\begin{equation}\label{eq:high_order1}
{a_i}(t+1)={f_i}\left( \sum_{j=1}^{k} {w}^{1}_{ji}{a_j}(t)+{w}^{2}_{ji}{a_j}(t-1)+\hdots+{w}^{n}_{ji}{a_j}(t-n+1)+w_{j0}\right)
\end{equation}
where  ${w}^{n}_{ij} \in [-1,1]$ stands for the casual relation originating from  $c_i$ and pointing to $c_j$ at time step $t-n+1$, while ${w}_{j0} \in [-1,1]$ is the bias weight related to the i-th node. 

As the above equation exhibits, the activation level of i-th concept at the moment $t+1$ depends on the activation values of all concepts at $\{t,t-1,\ldots,t-n+1\}$ moments, not only the activation values of the concepts at time $t$, during the consecutive iterations. 

$f_i$  is defined as the associated activation function for each concept. For instance if we consider sigmoid activation function, the activation function for each concept can be defined based on the steepness parameters $\gamma_i$ associated with $i$-th concept as follows:

\begin{equation}\label{eq:sigmoid}
f_i(x) = \frac{1}{e^{-\gamma_i{x}}+1}
\end{equation}

In vector term, the n-order FCM is replaced by the following formula:
 
\begin{equation}\label{eq:high_order_fcm}
{a}(t+1) = f\left( {W}^0+ {W}^1 {a}(t)
+\ldots+ {W}^{n} {a}(t-n+1) \right)
\end{equation}
where $f= [f_1 (t), f_2 (t),\dots, f_k (t)]^{T} $ is activation function vector where each member denotes associated activation function with the i-th concept. ${W}^{1}, {W}^{2}\dots,{W}^{k}$ are weight matrices to storing all weights of HFCM at $t,t-1\dots,t-k+1$ moment. $W^{0}= [w_{01}, w_{02},\dots,w_{0k} ]^T$ describes the bias vector of HFCM. ${a}(t+1),{a}(t),\dots,{a}(t-n+1)$ illustrates vectors of activation levels of all nodes in HFCM in $t+1,t,\dots,t-n+1$ moment. ${a}(t),{a}(t-1),\hdots,{a}(t-n+1)$ are the state vectors of HFCM in $t,t-1,\dots,t-n+1$,while ${a}(t+1)$ is the response vectors of HFCM for ${a}(t),{a}(t-1),\dots,{a}(t-n+1)$. It is necessary to mention that, according to the above formula, increasing the number of order leads to increasing the number of FCM parameters regarding to the size of weight matrices.

\subsection{Advanced Aspects}

\subsubsection{Extensions of FCMs}

According to the literature \cite{Papageorgiou2014FuzzyCM,Papageorgiou2012,nair2019generalised}, extensions of FCM have been designed to enhance the performance of the traditional FCM proposed by Kosko.  Fuzzy Grey Cognitive Maps (FGCM) \cite{salmeron2010modelling}, Intuitionistic Fuzzy Cognitive Maps (iFCM) \cite{Iakovidis2011intuinistic,Papageorgiou2012}, Belief-Degree-Distributed Fuzzy Cognitive Maps (BDD-FCMs) \cite{Ruan2011BDDFCM}, Rough Cognitive Maps (RCM) \cite{Chunying2011roughCM}, Dynamical Cognitive Networks (DCN) \cite{miao2001dynamical},  Evolutionary Fuzzy Cognitive Maps (E-FCM) \cite{cai2009creating}, Fuzzy Time Cognitive Maps (FTCM) \cite{park1995fuzzy},Dynamic Random Fuzzy Cognitive Maps (DRFCM) \cite{aguilar2003dynamic}, Rule-Based Fuzzy Cognitive Maps (RB-FCM) \cite{Carvalho2001rulebased}, Fuzzy Rules Incorporated in Fuzzy Cognitive Maps (FRI-FCM) \cite{Song2011FRIFCM} and Generalized FCM (GFCM) \cite{nair2019generalised} are some extended format of traditional FCM. 

Proposal of various FCM models was raised to cover FCM limitations including uncertainty problem or dynamic problems. For instance FGCM,iFCM,BDD-FCMs and RCM have been presented to deal with uncertainty problems. On the other side, DCN, E-FCM,FTCM and DRFCM have been developed to deal with dynamic problems. In 2019 the authors in \cite{nair2019generalised} proposed a new form of FCM called FGCM to deal both uncertainty and dynamic problems. Also, it provides a literature review of the conventional FCM approaches, and figures out the strengths and weaknesses of some important advances in FCMs to understand their potential in modeling complex qualitative systems. Since most of these models are employed in decision making domains, they are not elaborated in details in this review paper.

\subsubsection{Dynamic Properties of FCMs}
As discussed, FCMs produce a new state vector at each discrete time step in a repetitive process until the stop condition is met such as a maximum number of iteration or the system stabilizes. More precisely, the relational structure and inference mechanism of FCMs play critical roles in the dynamic behavior of FCMs to determine the state value at time $t+1$ by considering the previous state at time $t$. Based on this perspective, three possible scenarios are available to present the dynamic behavior of FCMs \cite{Gregor2013,felix2019review}. If the activation degree of the concepts remains stable after a finite number of time steps, then a fixed point is obtained with the convergence ability. Mathematically speaking, $\exists t_\alpha \in [1,2,\hdots,T-1]:{a}^{t+1}={a}^{t},\forall t\geq t_\alpha$ which means that ${a}^{t_\alpha}={a}^{t_\alpha+1}=\hdots={a}^{t}$. On the other side, in the limit cycle scenario, FCM generates dissimilar responses with the exception of a few states that are periodically produced. In other words, $\exists t_\alpha,P \in \{1,2,\hdots,T-1\}:{a}^{t+P}={a}^{t},\forall t\geq t_\alpha$ which is equivalent to  ${a}^{t_\alpha}={a}^{t_\alpha+P}={a}^{t_\alpha+2P}=\hdots={a}^{t_\alpha+jP}$ where $t_\alpha+jP\leq T$ such that $j\in \{1,2,\hdots,T-1\}$.Finally, in the Chaos scenario a different state vector generated at each iteration. Thus, the system is neither fixed nor cyclic.

Although updating the reasoning rule stops once a maximum number of iteration is reached, the output can be unreliable due to the lack of stability in chaotic or cyclic situations. It means that in order to better understanding the system behavior, stable responses are needed. So, convergence is often favorable in some scenarios including decision making and pattern classification. According to the literature \cite{felix2019review}, some factors including the method for updating concepts' values, the pattern encoded in the weight matrix and  non-decreasing activation function are the most convergence issues related to FCM-based systems. Accordingly, there exist some investigations around this subject such as  \cite{boutalis2009adaptive,kottas2007fuzzy,kottas2010fuzzy,kyriakarakos2012fuzzy,Gonzalo2014FCM} to improve convergence.

\subsubsection{FCMs against Artificial Neural Networks (ANNs)}
The relation between FCM and Artificial Neural Networks (ANNs) is one of the attractable issues and some studies have focused their main attention to compare FCMs with the other computational intelligence tools, in particular fuzzy rules knowledge-based systems and  Recurrent Neural fuzzy Networks (RNNs). In other words, FCMs can be expressed as interpretable RNNs based on their interpretability as the significant feature. From this perspective, FCMs consist of fuzzy logic elements during the knowledge engineering phase. In fact, the concepts of FCMs can be supposed as neural processing entities indicating that the activation degree of each neuron is defined through the value of the transformed weighted sum unit received from the connected neurons in the network.

Generally speaking, however, there are similarities among FCMs and AANs, there are some limitations and differences. From a structural point of view, an FCM is taken into account as a special type of RNN. Apart from topological similarity among FCMs and RNNs, they function differently. It is related to the knowledge interpretability of FCM as its super property. This feature enables FCM to be used successfully in modeling complex real-world problems. This means that, unlike traditional ANNs/RNNs, FCM neurons and the causal relations among them provide accurate interpretation of the system for humans. But in ANNs/RNNs, the neurons are considered as a computational unit without reflecting any clear meaning. In other words, an ANN/RNN performs its duty in a perfect manner but with limited interpretability because the presence of hidden neurons is neither interpretable nor explains why/how the solution is appropriate for the desired problem. On the other hand, in an FCM, each node has its meaning with a clear relation to other nodes. Besides, FCM can be regarded as an extension of Hopfield RNN \citep{Gregor2013} with a different structure while $w_{ij}=w_{ji}$ is needed to converge to a fixed point. Based on these differences, the modified ANN learning methods are used as FCM learning strategies and it is impossible to use them directly.  Therefore, as discussed, it is often said that there is a trade-off between the interpretability of FCM and the approximation ability of ANN. A few studies interpret FCM-based methods as ANNs \citep{tsadiras1999experimental,Tsadiras2008AFCMactivations,Papakostas2012hebbianFCM,napoles2016convergence,napoles2017learning}.

To encapsulate, like other fuzzy logic systems, learning has a basic role in designing FCM in practical applications. Although a variety of methods are introduced in the literature to adapt FCM, the close relation between FCMs and ANNs theories can provide a promising FCM learning strategy. Hence, the next section examines some important FCM learning methods proposed in the literature.
\section{Learning Algorithms}
\label{sec:learning}
The core contribution of FCM learning is to extract the weight matrix according to either expert intervention and/or the available historical data. In fact, learning algorithms aim to fine-tune FCMs. However diverse FCM learning methodologies are available in the literature, they are mostly classified in a triple class based on their fundamental learning pattern comprising Hebbian-based, population-based and hybrid techniques \cite{papageorgiou2011learning}. Therefore, this section concentrates specifically on examining and explaining the available methods in the literature.

\subsection{Hebbian-Based Learning Methods}

The particular aim of such unsupervised learning methods is to find weight matrices based on the domain experts' knowledge.  Various Hebbian-based learning algorithms exist in the literature and some of these relevant methods are introduced in the following. 

Differential Hebbian Learning (DHL) was suggested in \cite{Dickerson1993hebbian,Dickerson1994} based on the Hebbian theory \cite{Morris1999DOHT}; it is the first Hebbian-Based algorithm proposed by Dickerson and Kosko. In DHL, the weight matrices are modified when the value of corresponding concepts changes. Thus, the value of weights is updated repeatedly so that if there is no change in the value of concepts, the weight values remain the same in the next iteration. In other words, the learning process modifies the weight values until it reaches the desired state. The main drawback of this learning methodology is that the weight among a couple of concepts is updated by considering only the corresponding concepts and the effect from other concepts is ignored. Moreover, the order of data presentation plays a considerable role in the DHL method \cite{salmeron2019learning}. The weight is updated according to the following formula in this learning algorithm:
\begin{equation}\label{eq:DHL equation}
{w_{ij}}(t+1) = \left\{ \begin{array}{ccc}
        {w_{ij}(t)+\eta_i[\Delta{a_i}(t)\Delta{a_j}(t)-{w_{ij}(t)]}} & if & \Delta{a_i}(t) \neq 0  \\
        {w_{ij}(t)}& if & \Delta{a_i}(t)=0
    \end{array}\right.
\end{equation}
where $\eta_i$ is the learning factor, ${a_i}(t)$ is the activation value of concept $c_{i}$ and $\Delta{a_j}(t)={a_i}(t)-{a_i}(t-1)$ computes discrete changes along the time.

To handle aforementioned DHL issue, an improved version of DHL known as Balanced Differential Algorithm (BDA), was suggested in \cite{Huerga02abalanced}. Through this way, the change of all concepts at the same time step and with the same direction is considered in the weight updating process. This method improves on the limitation of DHL by considering the values of all the concepts that change at the same time as the weights are updated. Although DBA alleviates the limitation of DHL method, its application is restricted to just binary FCMs. Noteworthy that the following equation is employed to update the weights:
\begin{equation}\label{eq:BDA equation}
{w_{ij}}(t+1) = \left\{ \begin{array}{ccc}
        {w_{ij}}(t)+\eta_i[(\frac{\frac{\Delta{a_i}(t)}{\Delta{a_j}(t)}}{\sum_{k=1}^{n}
        \frac{\Delta{a_i}(t)}{\Delta{a_k}(t)}})-{w_{ij}}(t)] & if &  \Delta{a_i}(t)\Delta{a_j}(t)> 0 ,i \neq j \\
        {w_{ij}}(t)+\eta_i[(\frac{\frac{-\Delta{a_i}(t)}{\Delta{a_j}(t)}}{\sum_{k=1}^{n}
        \frac{\Delta{a_i}(t)}{\Delta{a_k}(t)}})-{w_{ij}}(t)] & if &  \Delta{a_i}(t)\Delta{a_j}(t)<0,i \neq j  \\
       
    \end{array}\right.
\end{equation}

In 2004 other two unsupervised Hebbian-based learning algorithms called Active Hebbian Learning (AHL) \cite{Papageorgiou2004active} and Nonlinear Hebbian Learning (NHL) \cite{papageorgiou2003fuzzy} were adopted while the learning of FCMs extremely depends on the expert’s intervention \cite{beena2011structural,Papageorgiou2005softcomputing,papakostas2011training}. 
In NHL method the range of concepts value, as well as the sign of edges, are specified by expert intervention while zero edges are not updated. Hence the main drawback of NHL is the construction of the initial graph which is recommended by experts. The initial graph structure elicited from the experts is remained during the learning process, thus, its physical interpretation is preserved. The stopping criterion is formed based on constraints imposed on nodes. The weights are adjusted when the stopping criteria are satisfied which include: (1) a close-enough solution to the desired response has been reached or (2) a fixed-point attractor has been identified. In this method the weights are updated using the following equation:

\begin{equation}\label{eq:NHL equation}
{w_{ij}}(t+1)=\left( {w_{ij}}(t)+\eta a_j(t)(a_i(t)-a_j(t).{w_{ij}}(t))\right)
\end{equation}

In AHL the desired set of concepts, an initial structure, and interconnections of concepts as well as the sequence of activation concepts are determined by expert knowledge. Unlike other methods, all weights are updated not only nonzero ones.  Thus, the weights in AHL are adjusted in seven steps iterative learning process to satisfy the predefined criteria. Furthermore, to prevent getting stuck in a local minimum, authors in \cite{li2004fuzzy} designed the Improved Nonlinear Hebbian Learning (INHL) technique in which a new term named impulse was added into the update rule. In this case, the rule to compute the weights can be summarized as follows:
\begin{equation}\label{eq:AHL equation}
{w_{ij}}(t+1)=\left(1-\gamma){w_{ij}}(t)+\eta a_i(t)[a_j(t)-a_j(t).w_{ij}(t)]\right.
\end{equation}

Data-Driven Nonlinear Hebbian Learning (DD-NHL) was introduced in \cite{stach2008data} as an improved NHL method. DD-NHL method relies on a similar learning principle as NHL, but the learning quality of DD-NHL is higher because of using historical data and decision concepts. Unlike other Hebbian methods, the initial weight matrix can be produced randomly in DD-NHL. Although the experimental results prove the super performance of the proposed method compared to NHL methods, its performance is poor in classification issues.

\subsection{Population-Based Learning Techniques}
In these supervised learning methods, the historical data is used in the shape of input-output pairs to train the model. In other words, we design the algorithm with the correct answer to each member of a dataset. In such problems, the algorithm tries to calculate the output for each new input by considering datasets into account. Therefore, the main goal is deriving the weight matrix to reflect the impact among the concepts by utilization of optimization algorithms to minimize error among the target and predicted responses. For this reason, these approaches can also be called Error-driven approaches \cite{felix2019review} and the experts are replaced by historical data. On the other hand, such learning methods aim to search the models to optimize the objective function, which is computationally expensive. Thus, automatic learning methodologies based on historical observations, unlike experts’ knowledge-based ones (conventional methods) with the lack of ability to model large size FCMs, become an important challenge. Accordingly, various types of evolutionary algorithms were introduced in the literature for training FCM and adjusting the optimal weight matrix.

The authors in \cite{Parsopoulos2002pso} introduced Particle Swarm Optimization (PSO) as a learning method of FCMs using historical data to determine and formulate the sub-optimal weight matrix of FCMs. The authors used the proposed method to minimize the fitness function to reach a desired final value for the FCMs with fixed architecture whereas determining the constraints depends on the human knowledge in this way.

Real-Coded Genetic Algorithm (RCGA) was exploited in \cite{stach2005genetic} as an FCM learning method to create FCM structure using the historical data in the frame of time series, which includes a single sequence of state vector values and without human intervention. The proposed RCGA is employed to minimize three various fitness functions to select the best one. Then selected function is used to perform experiments by considering a different number of concepts and FCM densities. The bigger the size of input data, the more precise learning is. On the other hand, the accuracy depends on the size of input data in the proposed RCGA and the learning performance degrades when increasing the size of the maps.

Multi-Objective Evolutionary FCMs is the other evolutionary learning algorithm proposed in \cite{mateou2005multi} based on GA to support multi-objective decision-making problems. This method is applied to search the optimal weight matrix which satisfies predefined activation levels among the participant nodes by the selection of the initial weights randomly.

Evolution strategy (ES) has been introduced in \cite{koulouriotis2001learning} as an efficient way to design and construct the FCMs. To ignore external intervention for fine-tuning of FCM parameters, more specifically in complex systems, the paper focused on ES as a robust and flexible training procedure. The proposed method, which is a composition of FCM and ES, is examined for potential implementation in FCM-based systems. In this algorithm, the learning process will stop when the optimal FCM structure is obtained.

In \cite{petalas2005fuzzy} a Memetic Particle Swarm Optimisation (MPSO) algorithm is suggested as FCM learning method to extract weight matrix by minimizing the fitness function to construct the desired system. The proposed MPSO is a combination of PSO as a global search algorithm and the Hooke and Jeeves (HJ) algorithm as a local search component. The final results have proved the outperformance of MPSO in comparison to PSO.

In \cite{ghazanfari2007comparing} Simulated Annealing (SA) algorithm was introduced as another metaheuristic FCM learning method to extract weight matrices from the input historical data without expert intervention. The comparison among the performance of SA and GA in this paper illustrates that with more nodes (complex FCMs), the GA algorithm will deteriorate the FCM performance meanwhile SA improves the learning quality by covering the GA limitation, as well as improving the speed of training for each nodes number. Thereby, the proposed method performs effectively for every map size. That is, the GA learning method is used for small FCM size while SA in large FCM size in this experiment.

Later an improved SA learning algorithm called Chaotic Simulated Annealing (CSA) was considered as FCM learning method in \cite{alizadeh2009learning}. According to the results, with more nodes, CSA outperforms SA with smaller learning error. Although the CSA performs well for every map size, the execution time is longer in comparison to SA. Thus, the authors introduced another new learning strategy termed as Tabu Search (TS) \cite{alizadeh2007learning} in which the quality of learning improved in comparison to GA. The TS algorithm generates smaller errors in comparison to GA as well as using fewer nodes. Besides, in some cases, the computational time of TS is less than GA especially for FCMs with small map sizes. To sum up, in terms of efficiency, the performance of CSA and TS is greater than conventional SA.

In addition to the above learning strategies, there exist other various types of population-based learning algorithms in the literature. For instance, \cite{salmeron2019learning} presents a novel learning technique namely asexual reproduction optimization (ARO) and Modified ARO (MARO). Also, it provides a comprehensive review of the available population-based learning methods. Game-based learning model was reported in \cite{Lue2009gamebased} by mixing FCM with game-based learning. Immune algorithm (IA) developed in \cite{lin2009immune} which can be utilized for designing and modeling complex systems, Big Bang-Big Crunch (BB-BC) in \cite{Yesli2010bigbang}, Ant Colony Optimization (ACO) as FCM learning algorithm in \cite{ding2011first}, Extended Great Deluge Algorithm (EGDA) in \cite{Baykasolu2011TrainingFC}, Artificial Bee Colony (ABC) algorithm in \cite{Yesil2013FuzzyCM}, Cultural Algorithm(CA) in \cite{Ahmadi2014cultural}, Imperialist Competitive Learning Algorithm (ICLA) in \cite{Ahmadi2015Imperialistcognitive} as a new robust, fast and accurate FCM learning method, Multi-objective optimization algorithm so-called MOEA-FCM in \cite{chi2015MOEA}, dynamic multi-agent genetic algorithm (DMAGA) proposed in \cite{liu2015dynamic}, evolutionary multi-tasking multi-objective memetic FCMs (MMMA-FCMs) learning algorithm adopted in \cite{Shen2019EvolutionaryMFCM}, Inactivation-based batch many-task  evolutionary algorithm (IBMTEA-FCM) in \cite{articleWang2021manytask} are some other examples of population-based FCM learning methodologies which have been introduced in the literature. 

\subsubsection{Fitness Functions of population-Based Learning Approaches}
With the goal of training FCMs, a myriad of fitness functions have been used to assess the performance of the designed FCM approaches based on the case studies. To be more precise, each chromosome's measurement of fitness function is done via one-step modeling of the relevant FCM. The output error among the actual responses of the system against the obtained state vectors, as FCM responses, is exploited to measure the fitness value. Therefore, the objective of fitness function applications is the assessment of the corresponding FCM through calculating the accumulated prediction error among the response values and real ones. The obtained error is averaged over the collection of nodes and the number of steps within the sequence of learning. Thus, this subsection reviews some objective functions available in the literature.

In \cite{koulouriotis2001learning} the authors proposed genetic learning method to optimize the objective function to set weights which transforms input to output vector as below.  $n$ and $k$ indicate the number of instances and concepts respectively.
\begin{equation}\label{eq:fitness_function_1}
 E(x)=\mathop{\sum_{k=1}^{k}\sum_{i=1}^{n}|a_{ki}-\tilde{{a}}_{ki}|}
\end{equation}
where $a_{ki}$ and $\tilde{{a}}_{ki}$ represent the current and expected responses respectively. 

The following equation unveils the Heaviside function (H) as the well-known fitness function introduced in \cite{Parsopoulos2002pso} using PSO as the learning approach. 

\begin{equation}\label{eq:fitness_function_2}
 E_{k}(x)=\sum_{i=1}^{n}H(L_{i}-{a}^{*}_{ik})|L_{i}-{a}^{*}_{ik}|+
 \sum_{i=1}^{n}H({a}^{*}_{ik}-U_{i})|{a}^{*}_{ik}-U_{i}|
\end{equation}
where ${a}^{*}_{ik}$ denotes the activation value of i-th decision concept, whereas $U_i$ and $L_i$ are upper and lower bounds respectively. It must be highlighted that this fitness function is not suitable for the gradient-based methods because it is not differentiable. This kind of fitness function is effectively used in PSO and memetic algorithms in \cite{Petalas2009impso,petalas2005fuzzy}.

In \cite{Papageorgiou2004Radiotherapy} investigators designed Differential Evolution (DE) learning method to minimize the objective function ($E(X)$ or $ F(W)$) defined as follows: 

\begin{equation}\label{eq:fitness_function_3}
 E(x)=\sum_{i=1}^{n}|L_{i}-{a}^{*}_{ik}|+|{a}^{*}_{ik}-U_i|
\end{equation}

The below equation represents the fitness function that was introduced in \cite{Stach2007Parallel,stach2005surveyFCM,stach2005genetic} while the parallel version of RCGA was applied to improve the scalability of the algorithm. 

\begin{equation}\label{eq:fitness_function_4}
 F(x)= \frac{1}{\alpha(error+1)}
\end{equation}
such that $\alpha$ indicates the fitness function coefficient and error is expressed through the following formula:

\begin{equation}\label{eq:error}
error=\frac{1}{n(k-1)}\mathop{\sum_{k=1}^{K}\sum_{i=1}^{n}|a_i(t)-{a}^{p}_i(t)|}
\end{equation}
where $n$ is the number of the historical data, $k$ is the number of concepts, $a_i$ and ${a}^{p}_i$ describe actual and predicted values respectively. 

In \cite{Stach2010divide} the researchers examined the divide and conquer approach for training large FCMs to optimize another fitness function, which was stressed as follows:
\begin{equation}\label{eq:fitness_function_5}
 F(x)=\frac{\alpha}{\mathop{\beta\sum_{t=1}^{T-1}\sum_{i=1}^{M}({a}^{t}_i-\tilde{a}^{t}_{i})}^2+1}
\end{equation}
$\alpha$, $\beta$ are positive scaling factors.

Besides the mentioned fitness functions, there exist very popular and useful fitness functions that were employed in papers which are described in the following \citep{salmeron2019learning,Shen2019EvolutionaryMFCM}.
\begin{enumerate}
    \item[]\textbf {In-sample-error}: This function is used to evaluate the learned model's ability to reconstruct the input data for learning. In other words, the model performance is calculated based on the pre-observed data via this criterion. In fact, the function measures the difference between the input data and the data generated by corresponding FCM from the initial state vector for the input data. Mathematically, it is explained as a normalized average absolute error between corresponding concept values at each iteration between the two sequences of state vectors based on the below equation \cite{koulouriotis2001learning,stach2005genetic,Ahmadi2015Imperialistcognitive,Stach2007Parallel,Stach2010divide,stach2008data,stach2012learning}:
    
    \begin{equation}\label{eq:in_sample_error}
    E(x)=\frac{1}{k(n-1)}\mathop{\sum_{t=1}^{N}\sum_{k=1}^{K}|a_{k}(t)-\hat{a}_{k}(t)|}
   \end{equation}
   
   where $a_k$ denotes the value of node $k$ at iteration $t$ in the input data, $\hat{a}_{k}$ is the value of node $k$ at iteration $t$ from simulation of the candidate FCM; $N$ the number of input data points; $K$ the number of nodes.
    
    \item[]\textbf {Out-of-sample-error}: computes the capabilities of generalization of the candidate FCM. To measure this criterion, both the input model and the candidate FCMs are simulated from randomly chosen initial state vectors. Therefore, the evaluation of the candidate FCM is based on new data that were not used for learning. This measure is normalized and expressed as an average error per concept per iteration similar to the in-sample error. The out-of-sample error is defined as: \cite{stach2005genetic,Stach2007Parallel,Stach2010divide,stach2012learning,Ahmadi2015Imperialistcognitive,chi2015MOEA,stach2010learning}.
    \begin{equation}\label{eq:out_of_sample_error}
   E(X)=\frac{1}{P(n-1)k}\mathop{\sum_{p=1}^{P}\sum_{t=1}^{N}\sum_{k=1}^{K}|{a}^{p}_{k}(t)-\hat{a}^{P}_{k}(t)|}
   \end{equation}
    where ${a}^{P}_{k}(t)$, $\hat{a}^{P}_{k}(t)$ are the values of node $k$ at iteration $t$ for data produced by input FCM and candidate  FCM respectively started from pth initial state vector. $N$ is the input data length, $K$ is total number of nodes and $P$ is the total number of different initial state vectors.
    
    \item[]\textbf{Error-l or Data-Error}: This function is mainly the variance or the average of the squared difference between two time series and is defined as follows \cite{Yesil2013FuzzyCM,Ahmadi2014cultural,Yesli2010bigbang}:
    \begin{equation}\label{eq:Error-l}
     E(X)=\frac{1}{k(n-1)P}\mathop{\sum_{P=1}^{P}\sum_{t=1}^{T-1}\sum_{k=1}^{K}({a}^{P}_{k}(t)-\hat{a}^{P}_{k}(t))^2}
    \end{equation}
    where ${a}^{P}_{k}$, $\hat{a}^{P}_{k}$ represent the k-th node value at iteration $t$ for the data generated via original FCM model and candidate FCM model respectively started from r-th initial state vector. Besides, $N$ is the length of data, $K$  is the nodes number and $P$ is the number of  randomly selected initial state vectors.
    
    \item[]\textbf{Model-Error}: This error compares the weight matrix of the trained FCM and the ground-truth weight matrix which is defined by the following formula: \cite{Shen2019EvolutionaryMFCM}

    \begin{equation}\label{eq:Model-Error}
    E(X)=\frac{1}{k^2}\mathop{\sum_{i=1}^{k}\sum_{j=1}^{k}|w_{ij}-{w}^{'}_{ij}|}
    \end{equation}
    
    where $w_{ij} $indicates the weight among node $i$ and node $j$in the real FCM whereas ${w}^{'}_{ij}$ is the weight among nodes i and j in the trained FCM and $k$ is the number of nodes.
\end{enumerate}

\subsubsection*{\textbf{Weaknesses and Strengths of The Proposed Learning Methods}}
As discussed, learning algorithms are employed in FCMs to extract/tune/adjust the weight matrix. With regards to the problem, based on the expert’s knowledge and historical data, the proper learning method used to construct an accurate FCM model. There are some strengths and limitations for both population-based and Hebbian-based methods.

Various learning methodologies are applied in different domains for FCM modeling, FCM time series prediction, FCM classification, FCM decision making, and FCM optimization as detailed in \cite{Papageorgiou2014FuzzyCM}. Based on the literature, population-based methods are used widely in a major proportion of the applications due to their lower simulation error, higher functionality, robustness, and generalization abilities \cite{papageorgiou2011learning}. Unlike the mentioned advantages, some generic limitations such as time-consuming, a large number of learning parameters, availability of historical data, a large number of learning processes, and convergence issues maybe hinder the application of learning approaches in some cases in this category. Hebbian-based methods, on the opposite side, are useful because of some features such as cheap cost of computing, ease of use, keeping signs of connection, and causal meaning of adjusted weights. Poor generalization, dependence on the experts, dependence on initial states and connections, low deviation from the initial weights can be taken into accounts as the main drawbacks for the Hebbian- based methods. Due to the mentioned analysis, it can be deduced that Hebbian-based algorithms are suitable for control problems \cite{Salmeron2013fuzzygrey}, while the population-based methods are applied more widely in areas such as time series forecasting, classification, simulating the chaotic behavior and virtual system, etc.\cite{stach2005genetic}.   

\subsection{Hybrid Methods}
As noted earlier, both Hebbian and population-based methods are confined in some cases due to their limitation. To alleviate the existing drawbacks and improve the efficiency, hybrid techniques which incorporate Hebbian and population methods were proposed as an alternative. In the sense that hybrid learning methods consider both features including the effectiveness of Hebbian learning and the global search ability of population-based methods to train FCMs. Indeed, the focus of hybrid methods is to updating/modifying the connection matrices extracted from the historical data and experts’ knowledge. Although hybrid algorithms can cope well with complex systems, according to the literature, little researches have been devoted to that method.

In\cite{Papageorgiou2005hybridFCM} a hybrid method was introduced which is a mixture of unsupervised learning methods, NHL algorithm, and differential evolution (DE) strategy to handle some FCMs drawbacks, improve the behavior of the FCMs dynamically and enhance the flexibility of FCMs. Ease of implementing, inexpensive computation, lower number of control parameters, comparable DE convergence properties as well as efficient handling of nonlinear, non-differentiable and multimodal fitness functions are the motivation factors of the investigated technique in this paper. The proposed NHL-DE algorithm is divided into two steps. Primarily the NHL method is employed to learn FCM then DE is used for FCM retraining. The goal of the first stage is to search for the appropriate weights while the responsibility of the second stage is recomputing the weights and minimizing the optimal fitness function. Noting that in this method the initial DE population directly depends on the performance of the first stage that is extracting a priori knowledge to incorporate in evolutionary computation is based on the good solutions in the first stage. Two termination conditions are considered in this method and the minimization process will be finished as the optimization criterion is satisfied in the second stage. The experimental results illustrate the high speed and efficiency of the model examined in three various FCM models.

Besides, the authors in \cite{Yanchun2008AnIF} introduced a new hybrid learning FCM method coupling NHL and RCGA algorithms. Exploiting RCGA and NHL as the key components of the algorithm improves the FCM's ability to extract data from the historical data and to explore the optimal weight matrices based on expert knowledge. This method benefits from both the effectiveness of the NHL learning method and the global search ability of the RCGA method. In the proposal method, RCGA is applied as an input data mining algorithm while the NHL is used for optimal weight refining. However the results are promising, the model is time-consuming as the size of the model increased.

In \cite{Ren2012hybridNHL} a hybrid method has been proposed using NHL and extended great deluge algorithm(EGDA). EGDA approach is similar to the SA training method with global search ability however it requires a lower number of parameters. Simplicity, high speed of convergence, and dealing with continuous concepts value are the main outstanding properties of the NHL algorithm. Taking the mentioned advantages of NHL and EGDA, the more accuracy of the hybrid method has.  Herein, after training FCM using the NHL algorithm, the outcome is fed to EGDA. In other words, the method has been constructed in two steps. In the first step, NHL is applied for FCM training to extract the optimal weight matrix. Afterward, the candidate objective function is optimized using the EGDA algorithm in the second step. 

\subsection{Other Methods}
There exist other methods proposed by investigators to construct or optimize FCM-based systems which do not belong to the triple above categories. 

\cite{konar2005reasoning} developed a novel unsupervised learning method. The weights of the directed edges are adapted from the transition to the Petri Net via the learning process. To analyze the dynamic behavior of the algorithm, the Hebbian-based learning algorithm was exploited considering natural decay in weights. After convergence of the learning algorithm, the network can be used to compute the beliefs of the desired propositions from the supplied beliefs of the axioms (places with no input arcs). The conditional nature of the algorithm in terms of stability allows the model to be used in complex decision making and learning. The other model was developed for knowledge refinement by adaptation of weights in a fuzzy Petri net using a different form of Hebbian learning. Also,  a combined fuzzy cognitive maps (FCMs)–petri nets (PN) approach
has been developed in \cite{Kyriakarakos2012petrinets} for the energy management of autonomous polygeneration microgrids. The PN is used as an activator in the
fuzzy cognitive map structure to enable different FCMs to be activated depending on the state of the microgrids.

Another technique based on Fuzzy Boolean Nets (FBNs) as a hybrid fuzzy neural technique adopted in \cite{Carvalho2007boolean} for training rule-based FCMs. In this method fuzzy Boolean nets are employed to extract qualitative fuzzy rules from crisp/quantitative data. Even though the main focus of this work is on the optimization and completion of Fuzzy Causal Rule Bases (FCRb), it can be generalized to all fuzzy rule-based. In other words, FBNs act like qualitative interpolators with perfect generalization ability. However, as the process generalizes, the performance of the method deteriorates due to the exponential increase in FBN size as the number of predecessors increases, even if it can be compensated by granularity employment in FBN’s internal memories.

In \cite{Zhang2017speech} a new FCM learning framework (e-FCM) proposed used for recognizing speech emotion. In this approach, the pleasure-arousal-dominance emotion scale is employed to measure the casual relations among emotions where the structure of the network is determined via certain mathematical derivations. The structure of e-FCM includes an input layer that collects data from speech features and an output layer composed of emotion classes.The proposed method is much faster and more accurate than traditional and population-based   methods such as GA particularly for large-scale FCMs.

Moreover, \cite{Mls2017InteractiveEO} outlined a learning method based on partial expert estimation and evolutionary algorithms to handle the lack of certainties in expert estimations of FCM weight matrices. This article hired a modification of Interactive Evolutionary Computing (IEC) for training and optimizing FCM termed as Interactive Evolutionary Optimization of Fuzzy Cognitive Maps (IEO-FCM).

Besides the aforementioned methods, the other FCM adaptation methods include gradient-based techniques. However these methods are used to minimize the cost function as a kind of error-driven approaches, they do not use metaheuristics. Therefore, they are classified in this learning group. A new FCM gradient-based learning method suggested in \citep{Madeiro2012Gradient} to improve the performance of population-based methods through their combination with local search approaches \citep{Madeiro2012Gradient}. Hence, this study evaluates the performance of the proposed automatic learning method which combines RCGA and DE with a gradient-based local search method. In fact, the mentioned method considers both abilities of exploitation and exploration in gradient procedures and evolutionary ones respectively. 

In \cite{Chi2014hybridmemetic} a composition of the memetic algorithm (MA) and artificial neural network (ANN) is adopted to learn large-scale FCMs, referred to as MA-NN-FCM. However MA is considered as a fast evolutionary-based method to find a set of regulated nodes, it suffers from the slow rating of weight matrix exploration. Therefore, a neural network employed to extract the weight matrix using the gradient descent approach. Experimental results prove the efficiency of the proposed method learning large-scale FCMs up to 100 nodes.

Based on the network topology of FCM, several FCMs learning methods inherited from ANN learning approaches to compute the weight matrix. \cite{Gregor2013gbsupervised} proposed a supervised gradient-based methodology based on delta rule and Backpropagation principle which originally developed for multi-layered networks. The extended version of the Backpropagation algorithm termed Backpropagation through time (BPTT) is applied in this investigation. Thus, three approaches including One-step Delta Rule, Every-step Delta Rule with Windowed BPTT and One-step Delta Rule with Windowed BPTT have been employed to measure the regression problem of FCM training.

The other automated FCM learning method introduced by \cite{chunmei2012cellular}. This approach exploits the evolutionary structure of cellular automata to train the weight matrix of FCM based on historical data consisting of one sequence of state vectors. Encoding the weight parameters are adopted via one-dimension cellular automata and a cell space is made by selecting the states of cellular in the range [0,1]. In this method the learning algorithm composed of 7 elements including where C-Cellular Automata code, E-fitness function, P-initialize Cellular Automata Configuration, M-initialize Cellular Automata Configuration size, S-selection, R-rule, T-stopping condition which can be defined in a 7-tuple (C, E, P, M, S, R, T). It is necessary to highlight that the employment of a mutation operator in the algorithm will improve the speed of convergence.

In \cite{Papageorgiou2016conceptreduction}, optimization technique was introduced to model complex systems containing a large number of concepts in the domain of decision making and management. The focus of this paper is to present a new FCM concept reduction approach and its application to develop less complex FCM which is easy to use. Accordingly, a clustering algorithm based on the fuzzy tolerance method was introduced as an FCMs reduction procedure by reducing the number of nodes and connections among them.  The concepts with the same behavior are classified into the same cluster. Afterward, the weight values are recalculated. The constructed FCM model with less complexity is capable to deal with the uncertainty in different scientific fields.

Deterministic learning of hybrid Fuzzy Cognitive Maps and network reduction approach proposed by \cite{napoles2020deterministic}. In this technique, FCMs are exploited to model dynamic systems, which are conditioned by several input variables that influence output ones. Thus, in this study expert knowledge combines with population-based knowledge mining as a hybrid method to create such a system. It means that the relation among input variables is determined by experts, while the extracted model from data provides the simulation of the output variable states. So, the scheme of the proposed model depends on both expert knowledge and historical data. As the other contribution, the authors introduced a very fast, deterministic (inverse method) and accurate learning rule to determine the weight matrix to define the system based on the Moore Penrose inverse, thus it does not require any parameter to be specified. Also, to guaranty that the learned weights are within the allowable range, a weight normalization approach has been considered. At last, a model proposed to distinguish irrelevant weights in the learned FCM network. As the model motivation, the absolute weight values have been taken into account as well as the concept activation values. Additionally, offset correction and weight correction methods were employed to tune FCM-based after removing a weight, however, the results verify that sigmoid slop calibration is more convenient than weight calibration in terms of efficacy. The reported MSE and processing time over 35 datasets indicate the superiority of the model compared with PSO, RCGA and DE.

In \cite{Feng2021TheLOentropy} a new straightforward, rapid, and robust learning method titled LEFCM proposed to tackle some limitations of convenient learning methods such as time-consuming dealing with large-scale FCMs ,lack of robustness when the experimental data contain noise as well as rarely weight distribution which affects the FCM performance. The crux of the investigated method is that the learning problem of FCM can be considered as a convex optimization problem with constraints that can be solved in polynomial time complexity by applying the gradient-based method. In the LEFCM method, the introduction of the least squares term into the convex optimization problem ensures the robustness of the well learned FCM. Furthermore, the performance of the generated FCM is improved because of the existing more reasonable weight distribution by considering entropy constraint in the convex optimization problem. Overall the proposed method is rapid and robust to learn large-scale FCMs up to 200 neurons, specifically, for learning FCMs using noisy data. 

Also, Gradient Residual Algorithm in \cite{zhang2011train}, Extreme Learning Machine in \cite{articleHuang2013elm}, Gradient-based search in \cite{gregor2013training}, Multi-step Gradient in \cite{inbook2015MultistepGradient} and $LASSO_{FCM}$ in \cite{Wu2016Robust}  belong to the last group of FCM learning methods, as can be seen from the Table \ref{Tab:FCM_learning}. 

This table summarizes the proposed FCM learning algorithms into different groups while population-based methods make up the major proportion of learning methods. More detailed look at the table reveals that some of the population-based algorithms have been introduced to deal with large scale problem. In other words, based on the paper \cite{articleWang2021manytask}, solving  the large-scale FCM learning problems can be solved exploiting three types of evolutionary-based algorithms including high-dimensional optimization technique \cite{liu2015dynamic,Yang2020LearningFC}, decomposition strategy \cite{chen2012GRN,YANG2019356} and search space reduction \cite{8071029zou2018,Liu2018InferringGR}.However, all of the these methods are  either ineffective or time-consuming to cope with the large-scale FCM learning problems due to the large search space \cite{articleWang2021manytask}. In order to enhance the speed and performance of the existing methods, a  random  inactivation-based  batch  multitasking  evolutionary  algorithm, IBMTEA-FCM was proposed in \cite{articleWang2021manytask} to  handle  large-scale  FCM  learning  problems. In this method, the  learning  of  local connections of nodes in a single FCM is modeled as a  many-task optimization (MaTO)  problem.

\begin{center}
    \begin{landscape}

\begin{longtable}[c]{|c|c|c|c|}
\hline
\textbf{Category} &\textbf{Learning Technique} & \textbf{Author} & \textbf{Node Num.}  \\ 
\hline

\multirow{5}{*}{Hebbian-based} & Differential Hebbian Learning (DHL) & \cite{Dickerson1993hebbian,Dickerson1994}& 10 \\ 

& Balance Differential Algorithm (BDA)& \cite{Huerga02abalanced}& \{5,7,10\}\\

& Nonlinear Hebbian Learning (NHL)& \cite{papageorgiou2003fuzzy}&5 \\

& Active Hebbian Learning (AHL)& \cite{Papageorgiou2004active}&8\\

&Data-Driven NHL & \cite{stach2008data}&\{5,10,20\}\\ \hline

\multirow{33}{*}{Population-based} &Particle Swarm Optimization(PSO) & \cite{Parsopoulos2002pso}&5 \\ 

& Genetic Algorithm(GA) & \cite{mateou2005multi}& 16\\

& Real Coded Genetic Algorithm (RCGA)& \cite{stach2005genetic}&\{4,6,8,10\} \\

&Parallel RCGA & \cite{Stach2007Parallel}&\{10,20,40,80\}\\

& Evolutionary Strategy (ES)& \cite{koulouriotis2001learning}& 6\\

& Memetic Particle Swarm Optimisation (MPSO) & \cite{petalas2005fuzzy}&18\\

& Simulated Annealing (SA)  & \cite{ghazanfari2007comparing}& \{2,3,...,15\}\\

& Chaotic Simulated Annealing (CSA)  & \cite{alizadeh2009learning}&14\\

& Tabu Search (TS) & \cite{alizadeh2007learning}&\{3,4,...,15\} \\
  
& Game based learning  & \cite{Lue2009gamebased}& 33\\

& Immune Algorithm (IA)  & \cite{lin2009immune}&26\\

& Differential Evolutionary (DE) & \cite{juszczuk2009learning}&5\\

& Big Bang-Big Crunch (BB-BC) & \cite{Yesli2010bigbang}& \{5,6,10\}\\

&divide and conquer RCGA & \cite{Stach2010divide}&40\\
& Ant Colony Optimization (ACO) & \cite{ding2011first}&40\\

&Extended Great Deluge Algorithm (EGDA)  & \cite{Baykasolu2011TrainingFC}& 6\\

&Sparse RCGA &\cite{stach2012learning}&40\\

& Decomposed ACOR based on ACO& \cite{chen2012GRN}&100\\

&Artificial Bee Colony (ABC)  & \cite{Yesil2013FuzzyCM}&13\\

& Cultural Algorithm (CA) & \cite{Ahmadi2014cultural}& 13\\

& Imperialist Competitive Learning Algorithm (ICLA) & \cite{Ahmadi2015Imperialistcognitive}& {\{5,7,8,10,13,20,24,30,40,50\}}\\

& Structure Optimisation Genetic Algorithm (SOGA)& \cite{poczketa2015learning}& 8\\
& Dynamic Multi-Agent Genetic Algorithm (DMAGA)  & \cite{liu2015dynamic}& \{5,10,20,40,100,200\}\\
&Decomposed RCGA with tournament selection&\cite{chen2015inferring2}&\{10,50,100,200,300\}\\

& Multi-objective Evolutionary Algorithm (MOEA-FCM)& \cite{chi2015MOEA}& \{5,10,20,40\} \\
&Decomposition-based DMAGA&\cite{WANG2021107441}&\{300,500\}\\
& Mutual Information based Two-phase Memetic Algorithm(MIMA-FCM)&\cite{8071029zou2018}& \{7,8,10,13,20,24,40,100,200,300,500\}\\
& Niching-based Multi-Modal Multi-Agent GA ($NMM_{MAGA}-FCM$) &\cite{YANG2019356}&\{5,10,20,40,100\}\\
& Asexual Reproduction Optimisation (ARO) and Modified ARO (MARO)  & \cite{salmeron2019learning}&\{5,6,7,10,24,100\} \\
& RCGA combined boosting strategy(RCGA-BFCMs)&\cite{yang2019learning}&\{10,20,40,100\}\\ 
&&&\\
&&&\\\hline
&  Multi-tasking Multi-objective Memetic FCMs (MMMA-FCMs)&  \cite{Shen2019EvolutionaryMFCM}& {\{10,20,40,100,200,400,600\}}\\
& MAGA based on the convergence error (MAGA-Convergence)&\cite{Yang2020LearningFC}&\{5,10,20,40,100,200\}\\ 
&Inactivation-based batch many-task  evolutionary  algorithm( IBMTEA-FCM) & \cite{articleWang2021manytask}&\{20,40,100,200,330,500\}\\
\hline

\multirow{5}{*}{Hybrid } & NHL+DE & \cite{Papageorgiou2005hybridFCM}&8\\

&RCGA +SA & \cite{ghazanfari2007comparing}&15\\ 

& NHL+RCGA&\cite{Yanchun2008AnIF}& 3  \\ 

&  NHL+EGDA& \cite{Ren2012hybridNHL}&3\\ 

&PSO+ACO & \cite{NAPOLES2014821}&25 \\\hline

\multirow{11}{*}{Other } &Gradient Residual Algorithm & \cite{zhang2011train}&7\\
&RCGA+ DE+a gradient-based method & \cite{Madeiro2012Gradient}& \{4,6,8,10,12,24,38\}\\
&Extreme learning machine &\cite{articleHuang2013elm} &6\\
& Gradient-based search&\cite{gregor2013training} & 20\\
& MA+Gradient descent & \cite{Chi2014hybridmemetic}& \{20,40,100\}\\
&Multi-step Gradient & \cite{inbook2015MultistepGradient}&8\\

&Extended Delta Rule & \cite{Rezaee2016multistage}& 34\\

&$LASSO_{FCM}$&\cite{Wu2016Robust}&\{10,20,40,100,200\}\\

& e-FCM & \cite{Zhang2017speech}& 5\\

&IEO-FCM & \cite{Mls2017InteractiveEO}&6\\ 
& Entropy-based method (LEFCM) & \cite{Feng2021TheLOentropy}& \{20,40,100,200\}\\

\hline
\caption{Proposed FCM learning techniques in the literature}
\label{Tab:FCM_learning}
\end{longtable}
\end{landscape}
\end{center}

\section{Time Series Forecasting using Fuzzy Cognitive Maps}
\label{sec:TSF using FCMs}

\subsection{Terminology of the problem}

Time series forecasting includes predicting the future observations based on the trained model on the historical observations. Thus, defining an accurate prediction model plays an essential role to make a better decisions in several fields for instance on engineering, medicine, economy, meteorology, etc.  In other words, let $Y(t)$ as given time series. The goal of time series forecasting models is to predict the next values of time series considering the prediction horizon ($H$) which can be divided into four categories including very short, short, medium and long term forecasting. On the other side, time series forecasting can be classified as univariate and multivariate models. In the case of univariate, only a single variable varies along the time. Accordingly, $ \hat{y}(t+H) = {M_u}({y}(t))$ is used to calculate the next values of time series with regards to the $H$ value where $M_u$ is forecasting model. In the case of multivariate time series, $\mathbf{\hat{Y}}(t+H)= {M_m} (\mathbf{Y}(t)) $ measures the next values for multivariate time series so that $\mathbf{Y}{(t)}= (y_1(t), \hdots, y_i{(t)}, \hdots, y_n{(t)})$ contains multiple variables that are varying over the time and $M_m$ is employed as the forecasting model. Besides, multivariate models can be grouped into Multiple Input Single Output (MISO) and Multiple Input Multiple Output (MIMO) for instance \citep{Silva2019Thesis} provides both MISO and MIMO fuzzy time series (FTS) forecasting models. In the MISO approaches, among numbers of variables, one of them is chosen as the endogenous (or target) variable while the others are considered as explanatory or exogenous variables. In contrast, in the MIMO approach, all variables have a chance to be forecasted or in the other words, all variables are target variables.

However, there is no perfect model to predict exact future values due the existence of the uncertainty and non-linearity associated with most of the real world phenomena. Accordingly, numerous time series forecasting methods have been presented in the literature to perform forecasting operation and estimate forecasted value $\hat{y}(t+H)$, from statistical methods (like ARMA, ARIMA, SARIMA, etc) to some new intelligent techniques (like LSTM, CNN, GRU, etc). Thereby, constructing the proper model is vital to minimize the prediction error between forecasted and real values. 

During recent decades, studies show remarkable advances of time series forecasting using FTS models based on some basic characteristics such as simplicity, readability, scaleability and high forecasting accuracy. Also, FCMs as weighted knowledge-based FTS techniques have found excellent applicability in the field of time series forecasting as a significant part of FCMs’ research in many areas. Therefore, the following subsections provide the general structure of FCM-based forecasting models as well as a review of some proposed FCM forecasting models in the literature.

\subsection{General FCM Model For Time Series Forecasting}

FCMs, as useful vehicles to represent time series in an easy and meaningful way, are a kind of interpretable recurrent neural networks, composed of neurons and weights. These weighted methods boost the forecasting accuracy by adding weights to the fuzzy rules compared with weightless ones. Therefore, designing proper FCM structure under given time series, calculating the FCM parameters (weight matrices) and reconstructing numerical values based on the FCM are essential considerations as they applied for time series modeling, time series dynamic behavior explanation and forecasting.

The forecasting model $M$ in FCM is related to the FCM-based network. It means that after mapping FCM neurons to time series variables, learning FCM is considered to extract the weights by testing within the prediction horizon. The basic issue in FCM-based time series forecasting models is formulating its structure, including the neurons and causal relations among pairs of them, based on the given time series. In time series, as a series of numeric data, the identification of concepts is not as easy as the traditional systems and needs to be explored \cite{Lu2014HFCMCMEANS}. For this reason, various studies on time series forecasting using FCMs, have been focused mainly on discovering the appropriate structure of FCM as well as defining the best learning algorithm to compute weight matrices to describe the dynamic behavior of the system. In simple terms, time series prediction by FCMs utilization consists of two stages. Firstly, designing the structure of FCMs by employing common successful strategies including  granularity \cite{Stach2008both}, membership values representation \cite{song2010fuzyyNNFCM}, fuzzy c-means clustering \cite{Lu2014HFCMCMEANS}. Furthermore, wavelet transformation and empirical mode decomposition (EMD)
are also proposed to identify FCM’s structure and enhance the forecasting performance \cite{LIU2020106105EMDHFCM,Shanchao2018WHFCM,gao2020robustemwt}. Secondly,  training  weight matrix using different learning methods as introduced earlier in section \ref{sec:learning}. Without high-quality weight matrices, the system will lose its interpretability, even with a crystal clear FCM reasoning \cite{vanhoenshoven2018fuzzy}. 

\subsection{Literature Review}
The main purpose of this section is to provide a review study of some of the relevant FCM based time series forecasting methods in the literature.

Apart from various partitioning methods proposed in the literature, fuzzy c-means clustering technique has been used widely to generate fuzzy sets (concepts) by dividing the Universe of Discourse($U$) into intervals with some overlaps. \cite{Homenda2014,Homenda2014TimeSM} are examples of univariate time series forecasting using FCM which fuzzy C-means clustering technique is applied to cluster all learning values of time series such that the number of clusters (denoted as concepts) is defined by the user and each cluster acts as a fuzzy set. Through this method, at any time, the membership degree indicates the value of the time series that belongs to the generated cluster. Meanwhile, the level of belongingness of the current value of time series $y(t)$ to the concept $c_i$ determined as activation value $A_i$. This procedure will be done for all concepts to reach the state vector. PSO used as a learning method to obtain weight matrix to optimize Mean Absolute Error (MSE) error among FCM responses and the targets.

The proposed method in \cite{Homenda2014} is based on the moving window technique, influenced by window size and widow step size, to extract the concepts. Then, a map composed of several layers is formed so that the obtained concepts in the previous step are located in each layer. Although it is an interpretable and clear FTS model, extracting appropriate concepts number is impossible. Thus, there is a trade-off between window size and accuracy in this model. In other words, the accuracy of this model depends significantly on the size of the window. The larger the window size, the bigger the map which creates some problems in relation to training, visualizing, and interpreting. On the other hand, the researchers in \cite{Homenda2014TimeSM} suggested a posteriori FCM simplification strategies, complexity reduction, by removing weak nodes and weights after evaluating their effectiveness in the map. Hereby it turned out that around $\frac{1}{6}$ of the edges and $\frac{1}{3}$ could be dropped without any substantial increase in the prediction error.

The authors in \cite{Lu2014Granules} unveiled a univariate time series forecasting method based on the coupling of FCMs and information granules employing  fuzzy c-means clustering technique to translate the original time series into granular time series. Thus, in the following, FCM is employed to describe obtained granular time series and accomplish forecasting tasks. PSO has been used as the learning algorithm in this method to extract the weight matrix among the concepts (granules). The proposed model is suitable for large-scale time series prediction because of its automatic structure. Noticeable that the more clusters, the higher accuracy, and the lower interpretability.

A feasible and effective time series forecasting framework based on fuzzy cognitive maps was investigated in \cite{lu2013linguistic} to predict the enrollments of the University of Alberta on the linguistic level. The proposed quantitative technique is a composition of the c-means clustering algorithm, FCM, and RCGA. The FCM structure formulated using c-means clustering which transforms original time series to FTS and extracts linguistic labels. Then, the constructed FCM is trained through the RCGA algorithm to represent the converted FTS and realize linguistic forecasting.

A new time series forecasting method was outlined in \cite{Lu2014HFCMCMEANS} by combining high order fuzzy cognitive map (HFCM) and fuzzy c-means clustering. The proposed model consists of two stages. In the first stage, HFCM model is constructed and the second stage exploits the generated HFCM prediction model to perform inference and prediction. In other words, firstly, a fuzzy c-means clustering algorithm is exploited to create information granules by translating original time series into granular time series and generate a structure of the HFCM model automatically. In the next, the PSO algorithm is implemented as a learning method to extract weight matrices and complete the HFCM prediction model. Eventually, the designed HFCM has been used to generate numeric prediction by doing inference in the granular space. Notice that the adopted method is driven through a couple of adaptable parameters: the number of clusters and orders. The best outcomes obtained when the number of clusters and order are 9 and 4 respectively. Based on the literature, it is kind of  Simple, legible with a high level of interpretability as well as a strong ability  to predict large scale time series.

In \cite{Homenda2014modeling} a new method has been implemented for time series forecasting where the lagged time series is mapped to the concepts. The method is a mixture of FCM reconstruction procedure with moving (sliding) window approach considered for both training and testing samples. In this method, The size of the map corresponds to the moving window size and it informs about the length of historical data, which produces time series model. It is worth mentioning that unlike other existing forecasting methods, fuzzification an defuzzification steps are not performed in this study and fuzzification has been replaced by min-max normalization in this paper. More precisely, time series modeling and forecasting depend on the moving window technique. PSO learning method is used to optimize objective function (MSE) for three real-time series data. Moreover, the performance of the model was evaluated in two cases: with and without the presence of the bias weights. Thus, this paper investigated the roles of map size and bias on the accuracy performance of the proposed method. Additionally, it is not equally good for all kinds of data and does not perform well for trend and seasonality data.

In \cite{Salmeron2016dynamicoptimisation} dynamic optimization of FCM (DFCM) was suggested to predict time series. The main goal of the DFCM method is to optimize not only the weights but also all elements of FCM as well as the whole learning process to reach more accuracy. Accordingly, the structure of the FCM together with the transformation function and its parameters as well as the length of the learning period are optimized. In this case, five different population-based algorithms such as GA, PSO, SA, ABC, and DE were applied to optimize all of the mentioned parameters. The proposed method is so competitive compared with other methods in the literature while is useful only for linear and stationary time series.

An extension of the time series forecasting method was designed in \cite{Stach2008both}. This two-level technique enables the model to carry out modeling and forecasting in both numerical and linguistic terms as the crux advantage by combining FCMs and granules. The outstanding characteristic of this candidate FCM corresponded to the benefits of the proposed RCGA learning technique. It must  be considered that the number of granular time series from the output of the fuzzification module represents the number of nodes in FCM-based modeling. Further, equal width of fuzzy sets and equal data frequency are used in this paper. The results indicate linguistic accuracy of the proposed method decreases as the number of fuzzy sets becomes higher, while at the same time, the numerical prediction accuracy increases. This shows that some trade-off exist between the quality of the numerical and linguistic predictions. Also, the tests show that the statistical characteristics of the input time series influence the quality of the results. A higher standard deviation of the input time series results in  a slightly worse accuracy of prediction.

Researchers in \cite{Wojciech2009comprative} published a comparative analysis of the evolutionary and adaptive learning method of FCM to evaluate the proper method for a special prediction problem. In other words, the performance of the proposed method evaluated by RCGA and DE as the evolutionary methods in comparison to DHL and balanced differential learning algorithm (BDA) as adaptive methods . The results demonstrate that the predictive capabilities of adaptive methods are not competitive with evolutionary algorithms. The predictive capabilities of the proposed method were examined by forecasting weather conditions in this study

The aim of the proposed research in \cite{Froelich2012ApplicationOE} is to create the FCM model for long-term prediction of prostate cancer based on an improved learning evolutionary learning method enabling FCM to better optimizing the fitness function for long-term prediction of multivariate time series. In other words, in this investigation, two types of FCM are designed for both short and long-term prediction by adding horizon parameter (H) to the prediction error. Therefore, two different FCM structures developed for $ H=1$ and $1\leq H\leq 7 $. The experimental results confirmed that obtained in-sample and out-of-sample prediction errors are much better for the second type by considering the propagation of errors that was occurred given $H > 1$. Noteworthy, the model complexity has been reduced by removing $|w_{ij}|<0.3$ .

The model elaborated in \cite{PAPAGEORGIOU201228} is another application of evolutionary learning FCM in which the main target is focused on the multi-step prediction of pulmonary infection based on the real clinical dataset. The experimental results confirm the good performance of the model for long-term forecasting and the capability of the proposed multi-step learning method to simulate fully both system dynamic nature and internal changes.

An unprecedented model for predicting chaotic time series designed in \cite{song2010fuzyyNNFCM,song2010FCMNNchoaic} implementing FCMs based on novel fuzzy neural network. In the contrary of the principal drawback of the conventional FCM models where determining the states of the investigated system and quantifying causalities mainly depends on the experts' knowledge, the proposed model equips the inference mechanism of original FCMs with the automatic identification of membership functions and quantification of causalities. Thereby, the construction of FCM can be modeled automatically from the data independently  with less intervention of experts. In this manner, the proposed fuzzy neural network composed of four layers. The first layer consists of input variables and each node represents a concept in the concerned system. The second layer performs fuzzification while each node represents the linguistic-term set of inputs, the third layer in which the nodes represent the output variables linguistic term, two tasks including the causalities among concepts in FCMs and defuzzification are carried out. Finally, in the last layer, the nodes represent non-fuzzy variables. The application of the proposed fuzzy neural network provides a crystal mathematical representation of the causalities and makes the inference process more understandable. Noting that Backpropagation-based learning algorithm is employed to adjust the relevant parameters by minimizing the RMSE as the objective function. Simple architecture and better forecasting accuracy confirmed the outperformance of the proposed method.

Despite several advantages of FCMs in solving decision-making problems due to their transparent and comprehensive nature, some FCM shortcomings limit its performance in the domain of time series forecasting and analysis. \cite{vanhoenshoven2018fuzzy} introduced a novel practical FCM method employing ARIMA to enabling FCM to predict multi-steps ahead of a fluctuating time series which keeps prediction accuracy, as well as keeping transparency of the weight matrix. In other words, to overcome the convergence issue, the updating rule adopted utilizing  moving average and weight amplification function. Also, It has applied a modified version of the sigmoid activation function to confine the activation values in the allowable range. However, the results confirm the advantage of the model in predicting multiple steps of a oscillating time series, some parameters like lower prediction accuracy as well as less transparency affect these achievements.

In another investigation, \cite{FROELICH20141319} proposed a new method  to alleviate the problem of forecasting multivariate interval-valued time series for the first time. For this purpose, FGCMs as an easy-to-interpret knowledge representation tool were designed as a non-linear predictive model. A genetic algorithm has been developed for training FGCM based on the historical data. In this model, the approximation of the time series is subjected to forecast instead of accurate numerical time series. As the experimental results indicated the better prediction performance is obtained as the prediction horizons are limited or for short-term prediction. For instance, it outperformed compared to ARIMA, VAR or naïve models for just up to three days prediction horizons.  Note that this investigation has used a sliding window concerning the both the size of training and testing windows and their effect on the final results. The model meets the better performance with the short prediction horizon, less data variability, and absence of trends in data.

The other approximate time series forecasting framework proposed in \cite{Froelich2016GranularFCM} which is a kind of new double-phased approach. In the first phase, the original time series is transformed to a sequence of granules to organize granular time series (GTS). In the second phase, the obtained granules from the first phase are clustered using fuzzy c-means clustering and the centers of obtained clusters are regarded  as the FCM concepts. Through this procedure, FCM is applied to forecast the concepts’ activation vectors.  Conceptually, the maximum activation level of the FCM concept is depicted the forecasted granule. Numerically, the forecasted granule is considered as a fuzzy set that is described in terms of its bounds and modal value. It should be reminded that in this research the GA has been defined as FCM learning method. The model is not suitable for random time series; however, it is the most proper for ones with a stable cyclic component as well as forecasting the changes of the amplitude.

In \cite{hajek2020intuitionistic} a new model is adopted for forecasting interval-valued time series mixing interval-valued intuitionistic fuzzy sets (IVI-FS) with a fuzzy cognitive map (FCM) trained via DE learning method. Also, this study has been focused on forecasting approximation of time series to create interval-valued time series (ITS). That is, the exact values are replaced by minimal and maximal values in the predefined periods. For this reason, an FCM-based model has been designed to predict ITS which is termed the Intuitionistic Fuzzy Grey Cognitive Map (IFGCM) using interval-valued intuitionistic fuzzy sets. The obtained results verify the satisfactory effect of the method compared by FCMs and FGCMs.

In \cite{Homenda2016clusteringFCM} an approach proposed for time series modeling and forecasting using clustering technique to design FCM in which the main focus is on structure or map node selection. In the proposed scheme, firstly, time series transformed into two-dimensional space of amplitude–change of amplitude as a basic and popular method. In other words, based on the \cite{Stach2008both} set of concepts in FCM are positioned in the two-dimensional space of amplitude and change of amplitude. Then, FCM clustering technique is used to create concepts $(k=3,4,5,..24)$. Cluster validity indexes are applied in the next step to evaluate the generated concepts. $40$ different validity indexes are exploited to evaluate the properties of the obtained concepts such as similarity of the points located in the same clusters, dissimilarity of the points belong to various clusters, both of them, or other features. Among all, just five cluster validity indexes performed well as top-five indexes. Step four includes the application of PSO as a learning algorithm for proposed FCM with $k=3,4,5, dots,24$ concepts to minimizing the MSE as the fitness function and evaluating the model performance. In the other words, the evaluation is centralized on selecting the best structure of FCMs using cluster validity indexes with the promising and minimum MSE value.

A combination of FCM with a multi-step supervised learning algorithm for time series prediction and monitoring was developed in \cite{poczketa2015monitoring}. Application of gradient model and Markov model of the gradient are used as a multi-step learning strategy for FCM in this investigation. On the other hand one-step gradient, multi-step gradient method, and Markov model of the gradient were employed to learn FCM and tune the relevant parameters by minimizing the forecasting error. Using the mentioned strategy provides the possibility of forecasting the next values according to the currently monitored values of the system. High flexibility, especially in small datasets, ease of usage, and cheap cost of implementation are the bold characteristics of the model. Most importantly though, less complex FCM produced through the SOGA training algorithm with semantic meaning in the provided relationships among concepts. The concepts of the obtained FCM is used as input for ANN with high accuracy.

The authors in \cite{Papageorgiou2015waterdemand} investigated a new FCM learning algorithm for the water demand problem. Thereby, the focal objective of this research is presenting the new SOGA learning method compared with other learning procedures including multi-step gradient model (MGM) and RCGA to optimize an objective function. In addition to investigate learning algorithms of FCM for multivariate time series modeling and prediction in this research, introducing a new FCM learning algorithm that finds the most important connections is considered as the other objective of this paper. However MSE, RMSE, MAPE, and MAE were used to evaluate the model's performance, it seems that the SOGA performs more efficiently regarding to the MSE as the accuracy metric of the model.  Further, based on the outcomes, the accuracy of summer demand prediction is better than winter demand. 

Furthermore, in \cite{Papageorgiou2016twostages} a novel ensemble time series forecasting technique was explored which combines FCM and artificial neural network (ANN). This multivariate method is based on the efficient capabilities of evolutionary fuzzy cognitive maps (FCMs) enhanced by structure optimization algorithms and artificial neural networks (ANNs) including two stages. In the first stage, an evolutionary FCM is made based on the SOGA in \cite{Papageorgiou2015waterdemand,zamora2014line}. Use of SOGA allows the construction of FCM model automatically from the historical data in which the most effective nodes (attributes) and weights are chosen to provide less complex and more efficient FCM-based model. In the next stage, the created FCM determines the inputs of ANN where the BP method with momentum or Levenberg-Marquardt (LM)algorithm IS exploited as ANN training strategies. Various kinds of accuracy metrics were employed to assess the model's performance using four various datasets. It is worth noting that the developed software tool ISEMK (intelligent expert system based on cognitive maps) used to evaluate the proposed method in all the experiments.

A novel two-stage forecasting technique introduced in \cite{poczeta2018implementing} which combines FCM and ANN to predict day-ahead gas consumption in Greece. In the first step, SOGA learning algorithm is employed to construct the FCM construction from the original time series.  In other words, SOGA is applied to extract the weight matrices as well as the model simplification by discarding the less effective concepts on the prediction accuracy. In the second stage, the selected concepts of the simplified FCM considered as inputs for ANN trained with the application of the back-propagation method with momentum and the Levenberg-Marquardt algorithm.

Moreover, another novel ensemble forecasting methodology developed by \cite{Papageorgiou2019FCMNNGAS} based on evolutionary FCMs, artificial neural networks (ANNs), and their hybrid structure (FCM-ANN).  In other words, the ensemble learning technique in this method combines various learning algorithms  including SOGA-based FCMs, RCGA-based FCMs, efficient and adaptive ANNs architectures, and a hybrid SOGA-FCM-ANN  recently proposed for time series forecasting structure proposed in \cite{Papageorgiou2016twostages} to solve the time series prediction problem of gas consumption in Greece.  In this research, two of the most popular ensemble methodologies including the simple average (AVG) and error-based (EB) \cite{lemke2010meta,makridakis1983averages} employed to evaluate the performance of individual predictors, the ensemble predictors, and their combination. The obtained results indicate the efficacy and efficiency of the defined method when compared with other autonomous methods like ANN, FCMs and LSTM.

In addition to the aforementioned options, to determine the FCM’s architecture and elevate the forecasting performance of FCMs, wavelet transformation and empirical mode decomposition (EMD) were recommended by some authors. In \cite{Shanchao2018WHFCM} a time series forecasting method has been adopted based on the synergy of high order FCM and wavelet transform (redundant Haar wavelet) which is termed as Wavelet-HFCM. The proposed method is useful to handle large-scale non-stationary time series while the original signal is decomposed into the multivariate time series by wavelet transform. Then, HFCM is exploited for modeling and forecasting the multivariate time series. Making more generalization, the proposed model was trained via ridge regression to solve optimization problems unlike the lasso regression procedure adopted for FCM leaning in \cite{Tsaih1998NLD,Martens2002volatility} . Based on the outcomes, the regularization factor and number of nodes have a considerable effect on the accuracy unlike the lower impact of the order.

An accurate and robust method to deal with non-stationary and large-scale time series was proposed in \cite{LIU2020106105EMDHFCM} which is based on the combination of empirical mode decomposition (EMD) as self-adaptive feature extraction technique and high-order FCMs (HFCMs), known as EMD-HFCM. Therefore, the EMD-HFCM is useful to handle some of the limitations of existing FCMs methods including low precision and sensitivity to hyperparameters. This novel method exploits EMD to create the set of stationary nodes of HFCM by extracting features from the original sequence. Then, a precise and efficient learning model based on Bayesian ridge regression, which is more robust than ridge regression, was employed to provide regular parameters from data.  The experimental results verify the excellent performance of the proposed EMD-HFCM on eight public time series datasets dealing with large-scale and non-stationary time series compared to other available models.

However \cite{Shanchao2018WHFCM} developed wavelet-HFCM method to handle large-scale non-stationary times series successfully, it suffers from some weaknesses relating to the wavelet transformation. Regarding this issue, a novel and robust forecasting technique proposed in \cite{gao2020robustemwt} based on the synergy of HFCM and empirical wavelet transformation (EWT) to boost the performance of conventional FCMs dealing with non-stationary and outliers. EWT used as a novel adaptive signal decomposition method with a significant effect on the model in analyzing non-stationary time series data. EWT is employed to decompose the original time series into different levels in the Fourier domain which captures information of different frequencies. Afterward, HFCM will be trained through a novel method based on $\epsilon$-support vector regression ($\epsilon$-SVR) which elevates its robustness against outliers.  In the end, the crisp value is obtained by dividing the summation of all the concept values of each node by two. The experimental results on eight publicly available time series show the superiority of the proposed model by comparing it with other models.

Although FCMs have a strong ability to apply for time series forecasting, the performance is limited due to some deficiencies using available feature extraction frameworks which influence the FCM’s prediction accuracy. In other words, in these methods, some features of the original time series will be lost when the original time series mapped to FTS. As instances \cite{Shanchao2018WHFCM} applied Harr wavelet to extract the features time series or fuzzy c-means in \cite{pedrycz2016design}. To deal with some limitations of proposed feature time series models, a novel and generalized feature time series extraction method is suggested bt \cite{yuan2020timekernelHFCM} merging kernel mapping and HFCM that has been inspired by the kernel methods and the support vector regression (SVR), referred to as kernel HFCM. Kernel mapping is defined to transfer the original one-dimensional time series into the multidimensional feature time series, and then key feature time series (KFTS) from the multidimensional feature time series are selected through the proposed feature selection algorithm to develop HFCM. In the next step, a fast HFCM learning algorithm based on ridge regression is applied to adopt the fuzzy relationship of the HFCM. Lastly, the predicted one-dimensional time series is generated from the feature time series by exploiting reverse kernel mapping.  However employing kernel mapping can help to capture the implicit patterns in the data, a suitable method for KFTS evaluation is still required to reduce the regression problem of the non-stationary time series and to improve the accuracy of the prediction.

A new time series forecasting strategy proposed in \cite{wu2019time} based on a sparse Autoencoder (SAE) and a high-order FCM (HFCM), named SAE-FCM to handle some time series forecasting pitfalls such as the inability of extracting good features of original time series, trapping in local minima and low prediction accuracy.  SAE has been exploited to dealing with the first limitation by extracting features from the original time series through an unsupervised training method. Then the second limitation is solved by a combination of outputs of SAE and HFCM  calculate the forecasted value. Finally, the batch gradient descent method (limited-memory Broyden-Fletcher-Goldfarb-Shanno (L-BFGS)), as fine-tuning algorithm inspired from deep learning, is used to update the weights of SAE-FCM and to improve the performance of SAE-FCM by removing the third limitation. The obtained results demonstrate the high accuracy performance of the proposed SAE-FCM. 

Another FCM-based time series forecasting strategy presented in \cite{vanhoenshoven2020pseudoinverse} using pseudoinverse learning model namely FCM-MP. Therefore, this study mainly has focused on developing a new time-efficient learning algorithm based on Moore-Penrose inverse to deal with some FCM learning limitations including time-consuming and poor accuracy of evolutionary and Hebbian-based methods respectively. High forecasting accuracy, cheap cost of computation, multiple-step-ahead multivariate forecasting, and lack of laborious adjustment of parameters are counted as the significant strengths of the proposed learning strategy. The experiment results on 41 different time series highlight the superiority of the mentioned technique considering the value of slope parameter of activation function $\lambda$ in the set  $\{1.0, 1.5, 2.0, 2.5, 3.0, 3.5, 4.0, 4.5, 5.0\}$.

The authors in \cite{wang2020deep} formulated a new extension of FCM for multivariate time series forecasting termed DEEP FCM (DFCM). In other expression, a deep neural network-based fuzzy cognitive map model was introduced to reach interpretable multivariate prediction. The proposed model combines the strength of interpretability of FCM with the strength of deep neural network by introducing the deep neural network models into the FCM knowledge-based models as the main solution for building an interpretable predictor strategy. DFCM as an extension of conventional FCM, modeled the nonlinear and non-monotonic effects among the concepts and unknown exogenous factors that have a latent influence on system dynamics via a fully connected neural network and a recurrent neural network (LSTM-based u-function) respectively. Furthermore, to calculate the strength of connection among a couple of concepts, the model equipped with a partial derivative-based approach to guarantee interpretability. An Alternate Function Gradient Descent (AFGD) approach based on Backpropagation (BP) also was exploited for parameter inference which enhanced its prediction ability compared to other standard FCM. 

The recent publication in \cite{a14030069} introduced a new time series modeling based on least square FCM termed as LSFCM. In this developed method, Fuzzy c-means clustering is employed to construct concepts of the FCM while the least square method is exploited to adopt the weight matrices from the given historical observation of time series. In contrast to other traditional FCM, the LSFCM model is a direct and one-time solution of matrix equation without repetitious stochastic searching. LSFCM model is a straightforward, robust, and rapid learning method, owing to its reliability and efficiency. In the first stage of this two-stage model, the least square method is exploited to form an FCM model. In the second step, the LSFCM is optimized through concept refinement by relocating the concept's positions to obtain the optimal concepts through PSO to improve the prediction accuracy. In this study, the slope parameter of sigmoid activation function $\lambda$ plays a significant role to keep the values of weights elements in the interval $[-1,1]$. It means that the proper weight matrix is obtained when $\lambda \geq \lambda_{1} $ and $\lambda_{1}= max\{|w_{ij}|\}$,$i,j=1,2,..,c$. The results confirm the efficiency and accuracy of the model compared with other conventional models considering various time series for the different values of concepts range from $\{2, 3,..., 20\}$ and $\lambda$ belongs to $\{1,5,10,50\}$.

A new FCM-based prediction model has been proposed in \cite{9494479fengpartitioningfcm} based on partitioning strategies. Firstly, fuzzy c-means clustering is employed to partition time series into several sub-sequence. Consequently, each partition has its corresponding sequences. Subsequently, the FCMs are constructed in terms of these sub-sequences respectively. Finally, the FCM models are merged by fuzzy rules. Therefore, this model has not been designed by modeling the whole data directly similar to other existing models in the literature. The constructed model performs well in numerical prediction and also has well interpretability. As the results indicates, the forecasting precision can not continuously improve by increasing the number of nodes. In reverse, the model is more accurate as the number of partitions increases. 

It is worth mentioning that Table \ref{tab:fcm_methods} summarizes a list of these proposed FCM-based time series forecasting strategies arranged according to the date of publication taking into account some important parameters such as learning methods, order number, presence of bias, and the number of nodes.

\begin{center}
    \begin{landscape}

\begin{longtable}[c]{|c|m{5cm}|c|c|c|c|c|} \hline
\textbf{Author}   & \textbf{Method} & ${\textbf{ Uni/Multi}}{\textbf{ variate}} $& ${\textbf{Learning}}{\textbf{ method}} $ &  \textbf{Order} & \textbf{Bias} & \textbf{Node Num.}      \\ \hline 

\cite{Stach2008both}    & FCM+information granule   &   Multivariate   &   RCGA   & $1$   &     without   & $9$ \\\hline
\cite{Wojciech2009comprative}   &\shortstack{Comparative study of adaptive\\and evolutionary FCM}  & Multivariate  &    {DHL, BDA, DE, RCGA}   & $1$ &   without & $4,5,8$ \\ \hline
\cite{PAPAGEORGIOU201228} &   Evolutionary based FCM  & Multivariate & 
\shortstack{Single-step RCGA\\, New multi-step learning method} & $1$  & without & $15$ 
\\ \hline
\cite{Froelich2012ApplicationOE}   & Evolutionary based FCM & Multivariate  &   RCGA  & $1$ &   withot & $ 6$ \\ \hline
\cite{lu2013linguistic} & {fuzzy C-means clustering+FCM} & Univariate  & RCGA  & $1$ & without & $5$ \\ \hline
\cite{Homenda2014}      & \shortstack{fuzzy C-means clustering +\\FCM+ moving window}     & Univariate      & PSO  & $1$             & without       & \shortstack{Node Num.=$3$,\\map size={3,6,..,27}}  \\ \hline
\cite{Homenda2014TimeSM}      & {fuzzy C-means clustering+FCM}     & Univariate     & PSO  & $1$             & without        & $27$\\ \hline
\cite{Lu2014Granules}  & \shortstack{fuzzy C-means clustering+\\ FCM+information granules}   & Univariate   &PSO &  1& with  & $\{3,4,5,6,7,8\}$ \\ \hline
\cite{Lu2014HFCMCMEANS}  & \shortstack {fuzzy C-means clustering\\+HFCM} & Univariate & PSO & \{1,2,3,4,5\} & with &     \{3,5,7,9,10\} \\ \hline
\cite{FROELICH20141319} & \shortstack{FGCM,FGCM/ARIMA,\\ FGCM/naive,FGCM/ES,\\ FGCM/VAR}  &  Multivariate &  GA & $1$ & without & $5$\\ \hline
\cite{Homenda2014modeling}  &  {Moving window+FCM} &  Univariate   & PSO   & $1$   & {with+without}   &   \{3,4,5,6,7,8,9,10,11,12\} \\ \hline

\cite{poczketa2015monitoring} & {Multi-step Learning Algorithm} & Multivariate & \shortstack{Gradient method and\\ Markov Model of Gradient} & $1$ & without & $22$\\ \hline

\cite{Papageorgiou2015waterdemand} & \shortstack{FCMs+
 learning algorithms \\based on gradient-based
\\and population-based methods} & Multivariate & MGM,RCGA,SOGA &
$1$ & without & $6$\\ \hline

\cite{Papageorgiou2016twostages} & FCM+ANN & Multivariate & \shortstack{RGGA,SOGA\\+BP with momentum and LM } & $1$ & without & \{8,9,14,26\}\\ \hline

\cite{Papageorgiou2016hybrid} & FCM+ANN & Multivariate & \shortstack{SOGA +BP with momentum\\ SOGA+ LM } & $1$ & without & $9$\\ \hline

\cite{Froelich2016GranularFCM} & \shortstack{fuzzy C-means clustering+\\granular FCM} & Univariate & GA & $1$ & without&  \{2,3,4,5,6\}\\ \hline

\cite{Homenda2016clusteringFCM} & \shortstack{New clustering based on cluster\\ validity indexes} & Multivariate & PSO & $1$ & without & $\{3,4,\hdots,24\}$\\ \hline

\cite{vanhoenshoven2018fuzzy} &  FCM+ARIMA  & Multivariate &  {Real-valued Genetic Algorithm} &  $1$ & without & $3$ \\ \hline

\cite{poczeta2018implementing} & FCM+ANN & Multivariate & {RCGA,SOGA+BP,LM} & $1$ & without & $6$ \\ \hline
\cite{Papageorgiou2019FCMNNGAS} & {Ensemble of Methods (AVG and EB) combining
FCMs and ANNs} & Multivariate & {SOGA,RCGA,BP} & $1$ & without& -\\ \hline

\cite{wu2019time}  & SAE+HFCM &  Univariate &  {Ridge regression+(L-BFGS)}  &  \{4,5,8,13\}  &   with   &  \{25,30,35,45\} \\  \hline

\cite{Shanchao2018WHFCM}    &     Wavelet+HFCM &       Multivariate &       Ridge regression & \{2,3,4,6,23\}  & with &   \{4,5,6,7,8\}\\ \hline

\cite{hajek2020intuitionistic} &  IFGCM & {Univariate+ Multivariate} & DE & 1 & without & 5 \\ \hline
\cite{LIU2020106105EMDHFCM}   &  EMD+HFCM &  Univariate &    {Bayesian ridge regression} 
&   \{2,6,9,11,22\}  & without &  \shortstack{self-adaptive\\ (depends on EMD)}\\ \hline

\cite{gao2020robustemwt} &     EWT+HFCM &     Univariate &     $\epsilon$-SVR  &  $\{12,48\}$   & without &  $3$ \\ \hline

\cite{yuan2020timekernelHFCM} & {Kernel mapping+ HFCM} & Univariate &  Ridge regression   &   \{2,3,4,5,6,7\} &  with &  \{2,3,4,5,6,7\}\\ \hline

\cite{vanhoenshoven2020pseudoinverse}  &  FCM-MP    &  Multivariate  &  {Moore-Penrose inverse} &  1  & with & \{2,3,4,5,6,10,14,50\}  \\ \hline

\cite{wang2020deep} & DFCM &  Multivariate   & {AFGD based on BP} & 1 & without & $\{6,9\}$ \\ \hline

\cite{Orang2020} & HFCM-FTS & Univariate & GA & 2 & {with+without} & $\{5,10,20\}$ \\ \hline

\cite{a14030069} & LSFCM &  Univariate & least squares & 1 & without & $\{2,3,..,20\}$ \\ \hline
\cite{9494479fengpartitioningfcm} & Partitioning strategies & Univariate & constrained least squares &1&
without& $\{3,5,7\}$ \\
\hline
\caption{Summary of the most relevant time series forecasting methods using FCM in the literature}
\label{tab:fcm_methods}
\end{longtable}
\end{landscape}
\end{center}

\section{Research opportunities and challenges}
\label{sec:challenges}

As the literature reviews and according to the Table \ref{tab:fcm_methods}, research on FCM-based forecasting models is continually increasing in recent years. Various univariate and multivariate FCM-based methods have been developed to enhance conventional FCM performance in terms of predictive accuracy and interpretation. Although the proposed models have achieved significant success in time series modeling and forecasting, there are some gaps and challenges ahead that are counted as follows. 

Firstly,  FCMs still comprise weakness in dealing with the nonstationarity and outliers even though \cite{gao2020robustemwt} has focused on this issue. Secondly, most of the models have been constructed to deal one step ahead forecasting and very limited number of papers focused on multiple steps ahead forecasting including \cite{vanhoenshoven2018fuzzy,PAPAGEORGIOU201228}. Thirdly, most FCMs implemented evolutionary algorithms as learning method which are very time consuming. So, the other significant challenge is to find a faster learning algorithm like least squares in \cite{a14030069}. The other challenges is to handle large-scale non-stationary multivariate time series which is considered in  \cite{Shanchao2018WHFCM}. In addition, scaleable learning methods, MIMO models, probability FCM forecasting can be considered as another future possibilities of FCM-based models.

\section{Conclusion}
\label{sec:conclusion}

This survey has been organized to provide an overview of recent developments on time series forecasting methods using FCMs and to explore potential future research opportunities. Also, this survey covers an introduction and revision on some corresponding properties/fundamentals of FCM (including the structure of FCMs and reasoning rules, high order FCMs, extensions of FCMs and dynamic properties of FCMs) and FCMs learning methodologies.

With respect to the FCM structure, the core attention of FCM-based approaches is basically focused on building FCM’s structure and extracting weight connections among the concepts. It means that common methods such as C-means clustering or information granules are employed to formulate the FCM structure and then learning algorithms are introduced to capture causal relation among a couple of nodes. Hebbian-based methods, population-based methods, their combinations, as well as ANN-based methods are the main category of FCM learning approaches, which, as mentioned, have advantages and disadvantages.  In addition, a variety of activation functions and reasoning rules were explored by researchers to generate exact and meaningful results by choosing the appropriate ones. 

FCMs as efficient member of soft computing society, have been widely used in time series forecasting domain to improve the accuracy as well as interpretability as a nonlinear predictive model. Time series forecasting applying FCMs includes two stages: designing a proper FCMs structure  and appropriate weight training techniques. Hence, different FCM models have been implemented to predict time series taking into account different numbers of concepts and orders and learning as well as various learning strategies. Since most FCMs used evolutionary learning algorithms, the recent remarkable challenges of FCMs is to find a fast training method such as least squares because  evolutionary-based learning methods are much time consuming. Although various univariate and multivariate methods have been presented to boost the conventional FCM particularly dealing with uncertainty in data, still exist some open challenges and future research possibilities as we discussed in section \ref{sec:challenges}.

\section*{Acknowledgements}

\bibliography{mybibfile}

\end{document}